\UseRawInputEncoding

\documentclass[journal]{IEEEtran}
\ifCLASSINFOpdf
\else
\fi

\usepackage{times}
\usepackage{epsfig}
\usepackage{graphicx}
\usepackage{amsmath}
\usepackage[ruled,vlined]{algorithm2e}
\usepackage{dsfont}
\usepackage{amssymb}
\usepackage{url}
\usepackage{amsfonts}
\usepackage{overpic}
\usepackage{multirow}
\usepackage{makecell}
\usepackage{booktabs}
\usepackage{lineno}
\usepackage{caption}
\usepackage{subcaption}
\usepackage{color}
\usepackage[table]{xcolor}

\begin{document}

\title{Bilaterally Slimmable Transformer for \\Elastic and Efficient Visual Question Answering}

\author{Zhou~Yu, \IEEEmembership{Member, IEEE},~
        Zitian~Jin,~
        Jun~Yu, \IEEEmembership{Member, IEEE},~
        Mingliang~Xu,~
        Hongbo~Wang,~
        Jianping~Fan
\thanks{This work was supported in part by the Zhejiang Provincial Natural Science Foundation of China under Grant LR22F020001, in part by the National Natural Science Foundation of China under Grants 62125201, 62072147, 62020106007 and 61836002, and in part by the Zhejiang Provincial Natural Science Foundation of China under Grant DT23F020007. (Corresponding authors: Jun Yu and Hongbo Wang.)}
\thanks{Z. Yu, J. Yu, and H. Wang are with School of Computer Science and Technology, Hangzhou Dianzi University, Hangzhou, 310018, China (e-mail: yuz@hdu.edu.cn; yujun@hdu.edu.cn; whongbo@hdu.edu.cn).}
\thanks{Z. Jin is with HDU-ITMO Joint Institute, Hangzhou Dianzi University, 310018, China (e-mail: jinzt@hdu.edu.cn).}
\thanks{M. Xu is with School of Computer and Artificial Intelligence£¬Zhengzhou University, 450001, China (e-mail: iexumingliang@zzu.edu.cn).}
\thanks{J. Fan is with AI Lab at Lenovo Research, 100094, China (e-mail: jfan1@ Lenovo.com).}
        }

\markboth{IEEE Transactions on Multimedia}%
{Yu \MakeLowercase{\textit{et al.}}: }
%





\maketitle

\begin{abstract}
Recent advances in Transformer architectures \cite{vaswani2017attention} have brought remarkable improvements to visual question answering (VQA). Nevertheless, Transformer-based VQA models are usually deep and wide to guarantee good performance, so they can only run on powerful GPU servers and cannot run on capacity-restricted platforms such as mobile phones. Therefore, it is desirable to learn an elastic VQA model that supports adaptive pruning at runtime to meet the efficiency constraints of different platforms. To this end, we present the bilaterally slimmable Transformer (BST), a general framework that can be seamlessly integrated into arbitrary Transformer-based VQA models to train a single model once and obtain various slimmed submodels of different widths and depths. To verify the effectiveness and generality of this method, we integrate the proposed BST framework with three typical Transformer-based VQA approaches, namely MCAN \cite{yu2019mcan}, UNITER \cite{chen2019uniter}, and CLIP-ViL \cite{shen2021much}, and conduct extensive experiments on two commonly-used benchmark datasets. In particular, one slimmed MCAN$_\mathsf{BST}$ submodel achieves comparable accuracy on VQA-v2, while being 0.38$\times$ smaller in model size and having 0.27$\times$ fewer FLOPs than the reference MCAN model. The smallest MCAN$_\mathsf{BST}$ submodel only has 9M parameters and 0.16G FLOPs during inference, making it possible to deploy it on a mobile device with less than 60 ms latency.
\end{abstract}


\section{Introduction}
\IEEEPARstart{T}hanks to recent progress on deep neural networks, machines are able to address complicated multimodal tasks that require a fine-grained understanding of both vision and language cues, such as image-text matching \cite{yu2020deep}\cite{dong2018cross}, visual captioning \cite{anderson2017up-down}, visual grounding \cite{yu2018mattnet}\cite{sun2022proposal}, and visual question answering (VQA) \cite{yu2019mcan}\cite{ouyang2021suppressing}. Among these tasks, VQA is challenging because it requires performing visual reasoning over multimodal data to predict an accurate answer.

\begin{figure}
    \centering
    \begin{subfigure}[h]{0.49\columnwidth}
        \includegraphics[width=\linewidth]{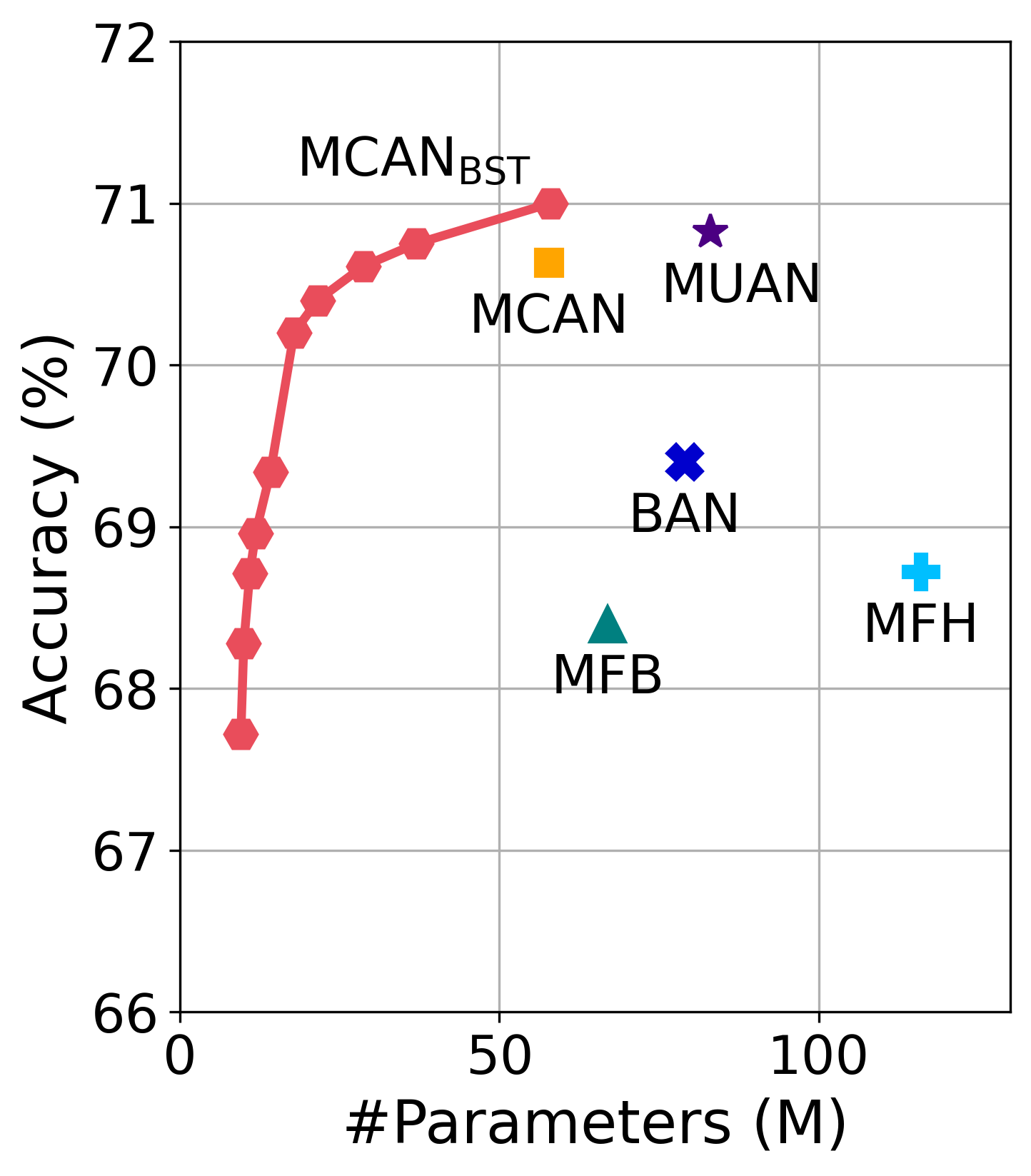}
        \caption{From-scratch training}\label{fig:sota_scartch}
    \end{subfigure}
    \begin{subfigure}[h]{0.49\columnwidth}
        \includegraphics[width=\linewidth]{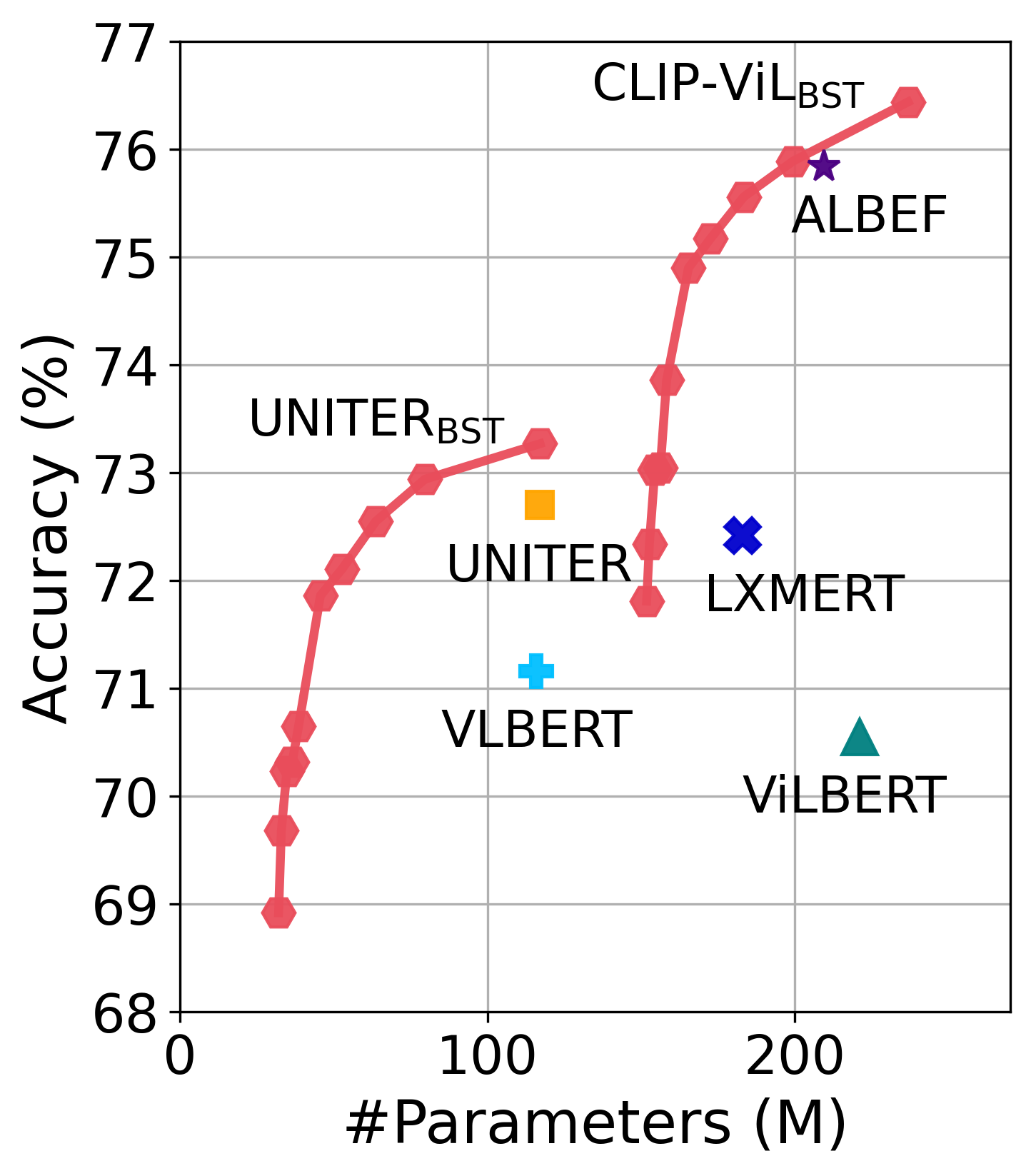}
        \caption{vision-language pretraining}\label{fig:sota_vlp}
    \end{subfigure}
    \caption{Accuracy \textit{vs}. number of model parameters on the VQA-v2 \texttt{test-dev} split to compare with the state-of-the-art methods. The methods are split into two classes depending on whether they are (a) trained from scratch \cite{yu2018beyond}\cite{kim2018bilinear}\cite{yu2019multimodal}\cite{yu2019mcan} or (b) pretrained with external data \cite{su2019vl}\cite{lu2019vilbert}\cite{tan2019lxmert}\cite{li2021align}\cite{chen2019uniter}. By integrating the BST framework with three typical VQA models, namely MCAN \cite{yu2019mcan}, UNITER \cite{chen2019uniter}, and CLIP-ViL \cite{shen2021much}, the resulted slimmable MCAN$_\texttt{BST}$, UNITER$_\texttt{BST}$ and CLIP-ViL$_\texttt{BST}$ models (marked with red lines) either outperform their counterparts with similar model sizes or achieve comparable performance with smaller model sizes, showing the \emph{efficiency} of our framework. Each red hexagon in the line represents a submodel sharing a portion of parameters of its full model (the rightmost one), showing the \emph{elasticity} of our framework.}
    \label{fig:bst_ind_example}
\end{figure}

Current state-of-the-art VQA approaches can be roughly categorized into two lines of research based on whether they are trained from scratch (\emph{e.g.}, MCAN \cite{yu2019mcan} and MUAN \cite{yu2019multimodal} in Fig. \ref{fig:sota_scartch}) or pretrained with external multimodal data (\emph{e.g.}, LXMERT \cite{tan2019lxmert}, UNITER \cite{chen2019uniter}, CLIP-ViL \cite{shen2021much}, and ALBEF \cite{li2021align} in Fig. \ref{fig:sota_vlp}). Although these two lines of research use different training paradigms, they share the same model architecture, \emph{i.e.}, Transformer \cite{vaswani2017attention}, which was initially proposed for language modeling and has since become the foundational architecture for the VQA task.

\captionsetup[subfigure]{font=small}

Despite the effectiveness of the Transformer-based VQA methods described above, they typically require large models (\emph{e.g.}, $\sim$200M) to guarantee good performance. This severely limits their applicability in capacity-constrained platforms with specific efficiency constraints (\emph{e.g.}, FLOPs and model size). 
To address this issue, a series of model compression strategies have been investigated to learn \emph{efficient} Transformer architectures, including low-rank decomposition \cite{ma2019tensorized}, weight sharing \cite{lan2019albert}\cite{dehghani2018universal}, model pruning \cite{cui2019fine}\cite{fan2019reducing}, and knowledge distillation \cite{tang2019distilling}\cite{sanh2019distilbert}\cite{sun2020mobilebert}\cite{liu2020fastbert}\cite{wang2020minilm}. However, these approaches focus on only one specific scenario and obtain one compact model. In practice, there are a wide variety of hardware platforms, \emph{e.g.}, GPUs, CPUs, and mobile devices. To meet the efficiency requirements of different platforms, compression methods need to redesign the model architectures and then retrain the models, which is both engineer-expensive and computation-expensive. This motivates us to devise an \emph{elastic-and-efficient} framework that supports training a single Transformer-based VQA model, and then adaptively pruning the model to fit different platforms \emph{without retraining}. Compared with the aforementioned model compression approaches, the introduced framework has the following two advantages: 1) it reduces the model design costs as the  submodels of different sizes can naturally meet the requirements of different platforms; 2) it also reduces the training cost as the model is trained once and adaptive slimming can be performed at inference time.

\begin{figure}
	\begin{center}
		\includegraphics[width=\columnwidth]{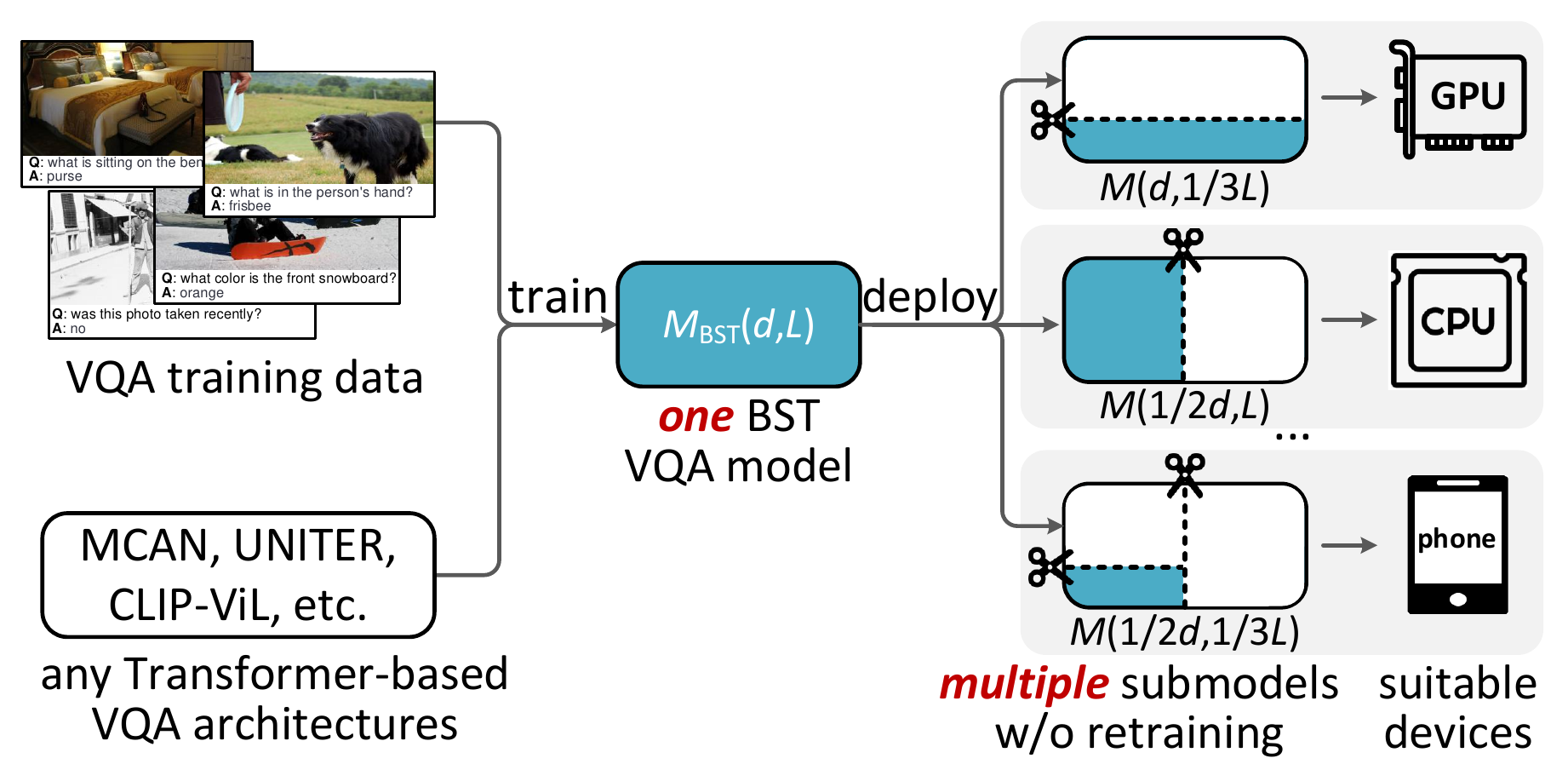}
		\vspace{-5pt}
		\caption{The schematic diagram of the proposed Bilaterally Slimmable Transformer (BST) framework. The BST framework is general enough to be seamlessly applied to arbitrary Transformer-based VQA models that supports training a \emph{single} BST-based VQA model only once and then pruning it to obtain \emph{multiple} efficient submodels to meet the requirements of different platforms at inference time.}
		\vspace{-10pt}
		\label{fig:scheme}
	\end{center}
\end{figure}

Inspired by the slimmable neural network that trains an elastic CNN model with multiple width multipliers to fit different efficiency constraints at runtime \cite{yu2018slimmable}, we present a bilaterally slimmable Transformer (BST) framework to support model slimming in both the width and depth directions of the Transformer, where the width is the hidden dimensionality of each layer and the depth is the number of layers. As shown in Fig. \ref{fig:scheme}, the BST framework can generally be seamlessly integrated with arbitrary Transformer-based VQA models to support training a single model only once and then pruning it to obtain multiple efficient submodels of various widths and depths at inference time. It is worth noting that each resulting submodel directly inherits a specific portion of the parameters of the full model and does not require further model finetuning. We take three typical Transformer-based VQA approaches, \emph{i.e.}, MCAN \cite{yu2019mcan}, UNITER \cite{chen2019uniter}, and CLIP-ViL \cite{shen2021much}, as the reference models to incorporate into the BST framework, resulting in the slimmable MCAN$_\texttt{BST}$, UNITER$_\texttt{BST}$, and CLIP-ViL$_\texttt{BST}$ models, respectively. As shown in Fig. \ref{fig:bst_ind_example}, the resulting slimmable models either outperform their existing state-of-the-art counterparts at similar model sizes or achieve comparable performance at much smaller model sizes.

To the best of our knowledge, our study is the first attempt to explore efficient and elastic models for VQA. The most closely related studies to our work is the DynaBERT approach \cite{hou2020dynabert} and the RWSAN approach \cite{qin2022deep}. DynaBERT also investigates slimmable Transformer architectures. In contrast to our BST framework, which supports arbitrary Transformer-based architectures for VQA, DynaBERT focuses on pretrained BERT model for NLP tasks. In terms of methodology, our BST is different from DynaBERT in terms of the slimming strategy and training strategy, resulting in better model performance and less training time. RWSAN investigates lightweight VQA models by introducing residual weight-sharing attention (RWSA) layers, resulting in a VQA model with many fewer parameters. In contrast to our BST framework, RWSAN cannot reduce computational costs and fit the efficiency constraints of different platforms adaptively. 

Our main contributions are summarized as follows:
\begin{itemize}
	\item Regarding the motivation, orthogonal to the pursuit of model accuracy, we introduce a new direction to the VQA research to learn an \emph{efficient and elastic} model once and obtain various efficient submodels that can be adaptively fit different platforms.
	\item Regarding the methodology, we present a general bilaterally slimmable Transformer (BST) framework which can transform any Transformer-based VQA model into a slimmable model to adjust the width and depth at runtime. Compared with DynaBERT \cite{hou2020dynabert} which introduces a slimmable BERT model for NLP tasks, our BST framework differs in terms of slimming strategies and training strategy.
	\item Regarding effectiveness and generality, we integrate the BST framework with three typical VQA models and conduct extensive experiments on two commonly used VQA datasets. The results show that the slimmed submodels either outperform the state-of-the-art approaches with similar model sizes or achieve comparable performance with smaller model sizes.
	\item Regarding practicability, this study is the first attempt to explore efficient VQA models on different hardware platforms including mobile devices. In particular. our smallest submodel can run on a non-latest mobile device with less than 60 ms latency, showing its potential in a wide range of applications such as robotics, automatic driving, and assistance for visually impaired people.
\end{itemize}

\section{Related Work}
In this section, we first briefly review previous studies on VQA, especially the approaches with Transformer architectures. After that, we discuss related work on efficient neural networks and slimmable neural networks.

\noindent\textbf{Visual Question Answering (VQA).} The VQA task, which aims to answer a free-form question in natural language with respect to a given image, has attracted increasing interest over the last few years. The core of VQA lies in two lines of research, namely multimodal fusion and attention learning. For multimodal fusion, early methods used linear models with elementwise summation or multiplication to fuse features from different modalities \cite{zhou2015simple}\cite{antol2015vqa}. To better characterize the second-order interactions between multimodal features, Fukui \emph{et al.} \cite{fukui2016multimodal}, Kim \emph{et al.} \cite{kim2016hadamard}, and Yu \emph{et al.} \cite{yu2017mfb} devised different multimodal bilinear pooling models. For attention learning, question-guided visual attention over image regions became a standard component of many early VQA approaches \cite{yang2016stacked}\cite{yu2020reasoning}. Yang \emph{et al.} proposed a stacked attention network to iteratively learn visual attention on different levels \cite{yang2016stacked}. More recently, co-attention models that consider both textual and visual attention have been proposed. Lu \emph{et al.} introduced a hierarchical co-attention learning paradigm to learn image attention and question attention iteratively \cite{lu2016hierarchical}. Yu \emph{et al.} decoupled the co-attention learning into a question self-attention stage and a question-conditioned visual attention stage and optimized the two stages in an end-to-end manner \cite{yu2017mfb}. The aforementioned co-attention models are \emph{coarse-grained} in that they neglect the multimodal interactions at a fine-grained level (\emph{i.e.}, word-region pairs). To this end, Nguyen \emph{et al.} \cite{nguyen2018improved}, Kim \emph{et al.} \cite{kim2018bilinear}, and Liu \emph{et al.} \cite{liu2020visual} introduced dense co-attention models that establish dense interactions among word-region pairs.

\noindent\textbf{Transformer-Based VQA.} The Transformer architecture is initially proposed for machine translation in the NLP community \cite{vaswani2017attention}. It consists of a sequence of self-attention modules to model complex and dense interactions within a group of input features. This architecture is general enough to be used in various unimodal tasks \cite{devlin2018bert}\cite{carion2020end}\cite{dosovitskiy2020image} and multimodal tasks in image \cite{yu2019mt}\cite{yu2019mcan} and video domains \cite{lei2021less}\cite{li2020hero}\cite{zhu2020actbert}. For the VQA task, Gao \emph{et al.} \cite{gao2019dynamic} and Yu \emph{et al.} \cite{yu2019mcan}\cite{yu2020deep} devised models based on the Transformer architecture and deliver new state-of-the-art performance on VQA benchmark datasets at the time of their publication. 
Besides these studies on general-purpose VQA, there is also a growing trend towards exploring more granular VQA tasks with specific reasoning skills, \emph{e.g.}, scene-text understanding \cite{yu2022knowledge}\cite{hu2020iterative}, casual reasoning \cite{liu2022causal}\cite{chen2020counterfactual}, knowledge utilization \cite{yang2022empirical}\cite{shao2023prophet}. 

More recently, a BERT model that integrates the Transformer architecture with a {self-supervised pretraining} paradigm has shown great success in a wide range of NLP tasks. Mirroring the success of BERT, recent studies have naturally extended its framework to the multimodal domain to perform vision-language pretraining (VLP) to learn generic multimodal representations \cite{lu2016hierarchical}\cite{tan2019lxmert}\cite{chen2019uniter}\cite{li2020oscar}. In particular, they first pretrain Transformer-based models on large image-text corpora to learn task-agnostic representations, and then finetune the models on downstream tasks such as VQA. Early VLP approaches designed different pretraining tasks to learn multimodal Transformers on top of preextracted region-based visual features \cite{lu2019vilbert}\cite{chen2019uniter}\cite{li2020oscar}\cite{yu2021ernie}\cite{cui2021rosita}. Motivated by the success of pretrained visual backbones, \emph{e.g.}, ViT \cite{dosovitskiy2020image} and CLIP \cite{radford2021learning}, recent VLP methods tend to exploit these visual backbones to obtain grid-based visual features and perform multimodal pretraining from raw image and language inputs in an end-to-end manner \cite{wangofa2022}\cite{li2021align}\cite{li2022blip}\cite{shen2021much}\cite{dou2022empirical}.
To summarize, Transformer-based approaches dominate the VQA task at present, due to their excellent capability for modeling the complex interactions among multimodal input features. However, Transformer-based VQA models are usually computationally expensive (\emph{i.e.}, have a large number of parameters and FLOPs), hindering their deployment on mobile devices with limited memory and computation consumption. This motivates us to explore efficient Transformer architectures for VQA. 

\noindent\textbf{Efficient Neural Networks.} There has been broad interest in building efficient neural networks in the literature. Existing approaches can be generally categorized as either compressing pretrained networks \cite{han2015deep}\cite{han2015learning}\cite{liu2017learning} or training efficient networks directly \cite{iandola2016squeezenet}\cite{howard2017mobilenets}\cite{zhang2018shufflenet}. The efficient neural networks above mainly focus on ConvNet architectures. Due to the popularity of Transformer in recent years, efficient Transformer architectures have been investigated in different respects, \emph{e.g.}, low-rank decomposition \cite{ma2019tensorized}, weight sharing \cite{lan2019albert}\cite{dehghani2018universal}, model pruning \cite{cui2019fine}\cite{fan2019reducing}, and knowledge distillation \cite{tang2019distilling}\cite{sanh2019distilbert}\cite{sun2020mobilebert}\cite{liu2020fastbert}\cite{wang2020minilm}. Low-rank decomposition methods decompose a full-rank parameter matrix into low-dimensional matrices while weight sharing approaches reuse one parameter matrix in different layers of the network. Model pruning methods aim to cut out redundant parameters in the model to obtain a smaller model, and knowledge distillation techniques aim to transfer knowledge from a large teacher model to a small student model. Despite the success of these approaches, their efficient models are dedicated to one specific scenario, and cannot adapt to different efficiency constraints or different hardware platforms at runtime.

\noindent\textbf{Slimmable Neural Networks.} Orthogonal to the approaches to efficient neural networks above, slimmable neural networks aim to design \emph{dynamic} models that can adaptively fit different efficiency constraints at runtime. Given a deep neural network, network slimming can be performed on both the depth and width dimensions. For depth slimming, the methods of Wu \emph{et al.} \cite{wu2018blockdrop}, Liu \emph{et al.} \cite{liu2017learning}, and Huang \emph{et al.} \cite{huang2016deep} learn controllers or gating modules to adaptively drop layers from deep ConvNets. For width slimming, Yu \emph{et al.} introduced a general framework for a family of ConvNets (\emph{e.g.}, ResNet \cite{he2015deep} or MobileNet \cite{howard2017mobilenets}) that supports a predefined set of width multipliers \cite{yu2018slimmable}. After that, they further improved the framework to support model slimming with arbitrary widths \cite{yu2019universally}. To take a further step, Cai \emph{et al.} introduced a once-for-all (OFA) method to support width and depth slimming simultaneously in a unified framework \cite{cai2019once}. All of the methods above are only for ConvNet architectures, and their strategies cannot be directly applied to Transformer architectures.

The most closely related study to our work is DynaBERT \cite{hou2020dynabert}, which also investigates slimmable Transformer architectures. In contrast to our BST framework, which supports arbitrary Transformer-based architectures for VQA, DynaBERT focuses on pretrained BERT model for NLP tasks. Regarding the methodology, our BST is different from DynaBERT in terms of slimming strategy and training strategy, obtaining significant advantages in terms of a higher compression ratio and less training time.

\section{Bilaterally Slimmable Transformer (BST)}
In this section, we describe the bilaterally slimmable Transformer (BST) framework in detail. Before presenting the BST framework, we first revisit the core components of the Transformer architecture \cite{vaswani2017attention}. Then, we introduce the BST framework, including the slimming strategies for width and depth. We take three typical VQA models, MCAN \cite{yu2019mcan}, UNITER \cite{chen2019uniter}, and CLIP-ViL\cite{shen2021much} as examples and integrate then with the proposed BST framework. Without loss of generality, our BST framework can be applied to arbitrary VQA models of Transformer-based architectures. After that, we introduce the BST training strategy which consists of submodel architecture selection strategy and a knowledge distillation training stage. Finally, we provide in-depth comparisons to DynaBERT \cite{hou2020dynabert} in terms of methodology. 

\subsection{Preliminaries}
Transformer is a multi-layer network in which each layer consists of the multi-head attention (MHA) and feed-forward network (FFN) modules \cite{vaswani2017attention}.
\\
\noindent\textbf{Multi-Head Attention.} Denote $m$ query features and $n$ key-value paired features as $Q\in\mathbb{R}^{m\times D}$, $K\in\mathbb{R}^{n\times D}$, and $V\in\mathbb{R}^{n\times D}$, where $D$ is the hidden dimensionality of these features. The multi-head attention module calculates the attended features $F\in\mathbb{R}^{m\times D}$ by using $H$ paralleled attention functions as follows:
\begin{align}\label{eq:mha}
F = \texttt{MHA}(Q,K,V)&=[\mathrm{head}_1,\mathrm{head}_2,...,\mathrm{head}_H]W^o \nonumber \\
 \mathrm{head}_j&=\texttt{ATT}(QW_j^Q,KW_j^K,VW_j^V)
\end{align}
where $W_j^Q, W_j^K, W_j^V \in\mathbb{R}^{D \times D_H}$ are the projection matrices for the $j$-th head, and $D_H$ is the dimensionality of the features from each head. $W^o\in\mathbb{R}^{H\cdot D_H \times D}$ is the projection matrix used to aggregate the output features from different heads. We set $D_H=D/H$ so that the model sizes remain constant as $H$ varies. The attention function for each head is defined as the scaled dot-product of the query with all keys:
\begin{equation}\label{eq:sdp}
\texttt{ATT}(Q,K,V)=\mathrm{softmax}(\frac{QK^T}{\sqrt{D_H}})V
\end{equation}
which calculates the scaled dot-product of each query with all keys to obtain the attention weights, and then performs weighted summation over the values.
\\
\noindent\textbf{Feed-Forward Network.} The feed-forward network module is a two-layer MLP model applied to the output features of the MHA module to perform a pointwise nonlinear transformation. Given input features $X\in\mathbb{R}^{n\times D}$, the transformed features $F\in\mathbb{R}^{n\times D}$ are obtained as follows:
\begin{equation}\label{eq:ffn}
F = \texttt{FFN}(X)=\mathrm{ReLU}(XW_1+b_1)W_2^T + b_2
\end{equation}
where $W_1, W_2\in\mathbb{R}^{D\times 4D}$.
\vspace{5pt}
\\
\noindent\textbf{Transformer Layer.} A typical Transformer layer usually consists of a MHA module and an FFN module as follows:
\begin{equation}
\begin{aligned}\label{eq:transformer_layer}
{F} &= \texttt{Transformer-layer}(X)\\
 &= \texttt{LN}(\texttt{FFN}(\hat{F})+\hat{F}) \\
 \hat{F} &= \texttt{LN}(\texttt{MHA}(X,X,X)+X)
\end{aligned}
\end{equation}
where residual connection \cite{he2015deep} and layer normalization (LN) \cite{ba2016layer} are applied after the MHA and FFN modules. The LN module takes a $D$-dimensional feature $x\in X$ as its input and performs normalization as follows to obtain the output feature:
\begin{equation}\label{eq:ln}
y=\texttt{LN}(x)=\gamma\frac{x-\mu}{\sqrt{\sigma^2-\epsilon}}+\beta
\end{equation}
where $\mu,\sigma^2\in\mathbb{R}$ are the mean and variance calculated on $x$. $\gamma, \beta\in\mathbb{R}^D$ are the learnable parameters of the scale and shift terms, respectively.
\\
\noindent\textbf{Transformer Architectures.} Depending on the composition strategies of the Transformer layers above, existing Transformer architectures can be categorized into three classes, namely encoders \cite{devlin2018bert}\cite{liu2019roberta}, decoders \cite{radford2018improving}\cite{radford2019language}, and encoder-decoders \cite{vaswani2017attention}\cite{raffel2019exploring}.

\begin{figure*}
	\centering
	\begin{subfigure}[h]{0.61\textwidth}
		\includegraphics[width=\linewidth]{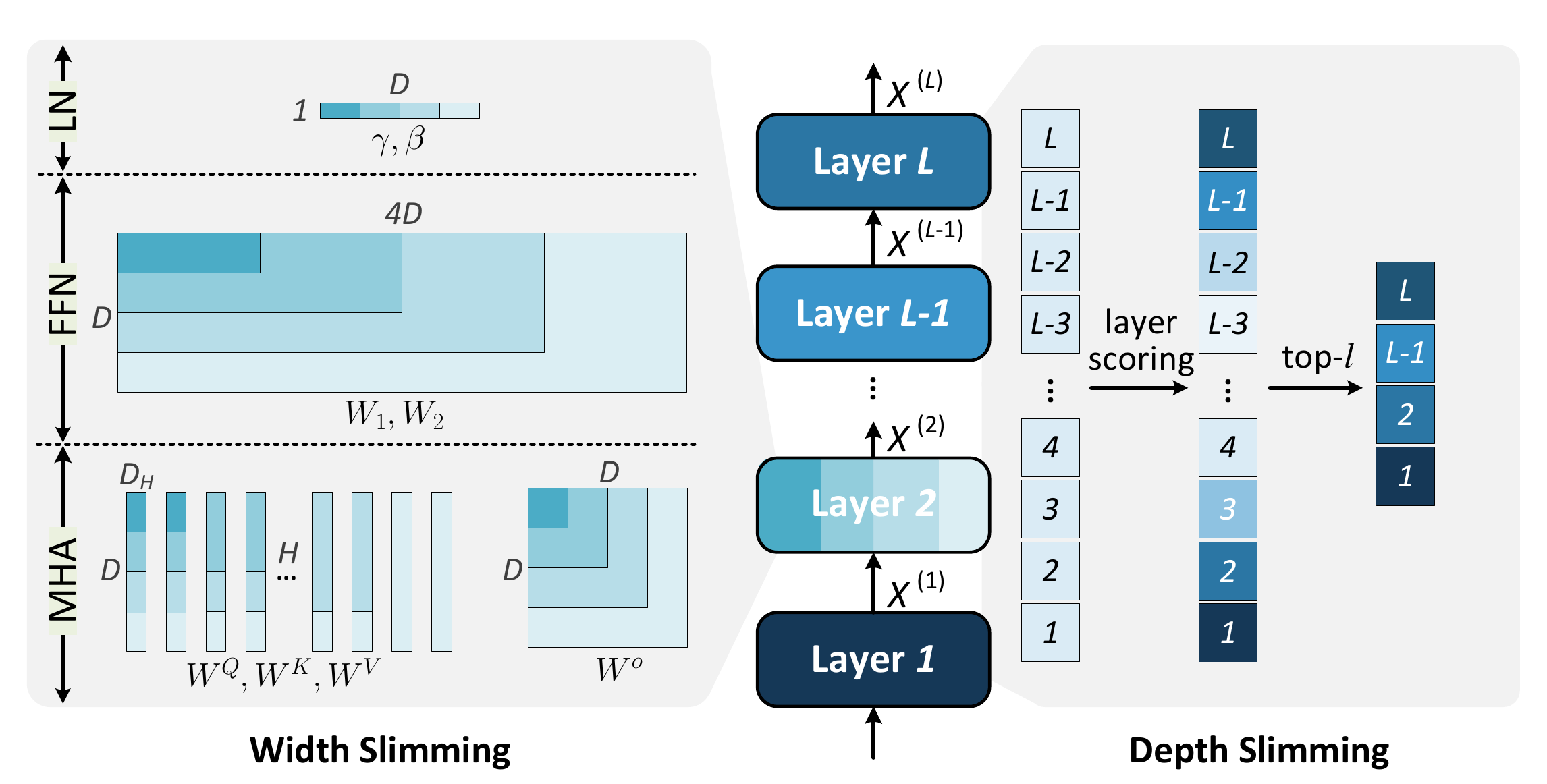}
		\caption{The general BST framework supports both width and depth slimming}\label{fig:slim_strategy}
	\end{subfigure}
	\quad
	\begin{subfigure}[h]{0.35\textwidth}
		\includegraphics[width=\linewidth]{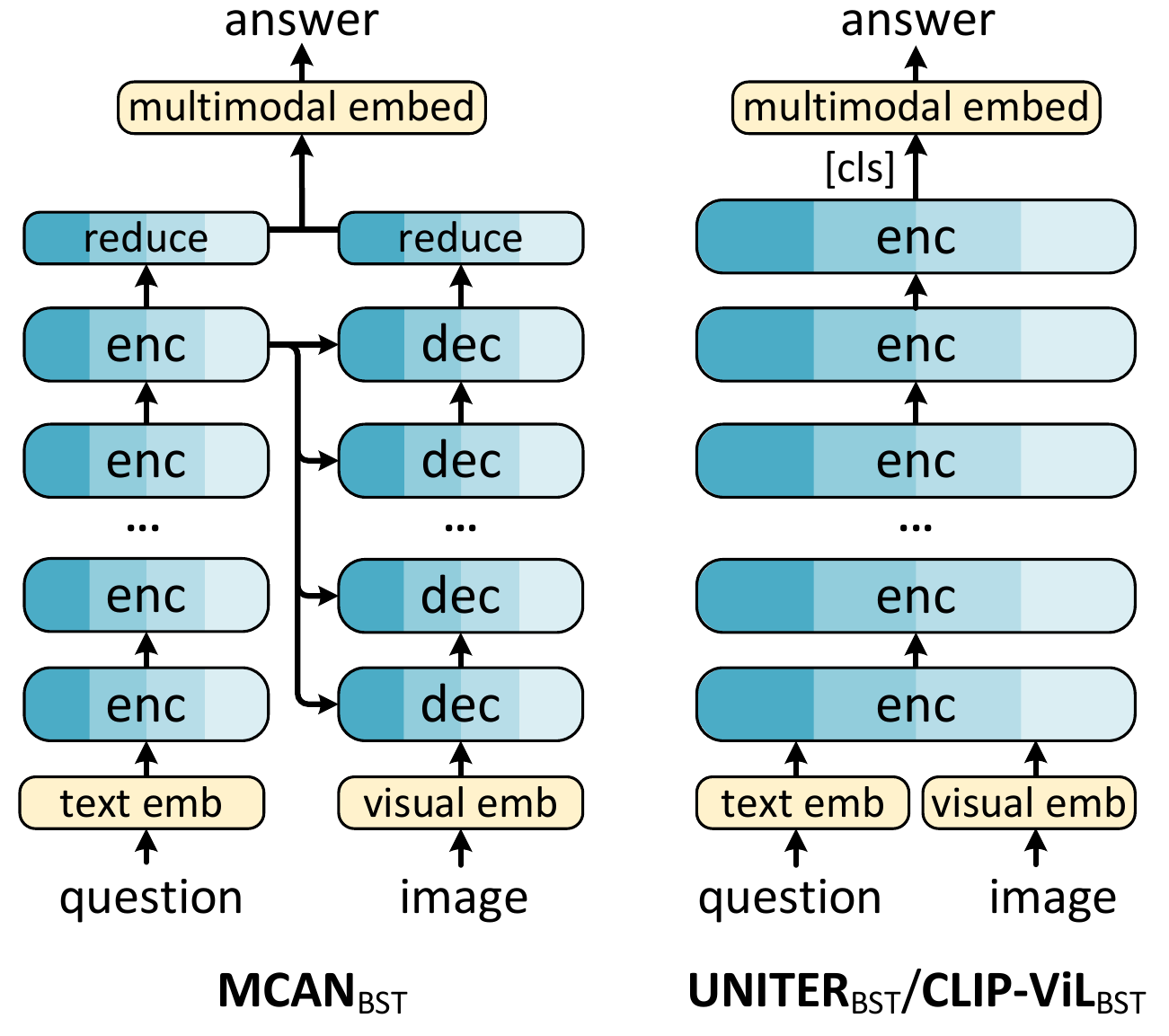}
		\caption{Integration of BST into existing VQA models}\label{fig:bst_vqa}
	\end{subfigure}
	\caption{(a) The overall diagram of the proposed Bilaterally Slimmable Transformer (BST) framework, which consists of width slimming and depth slimming. (b) The integration of the BST framework and three typical Transformer-based VQA models, namely MCAN \cite{yu2019mcan}, UNITER \cite{chen2019uniter}, and CLIP-ViL \cite{shen2021much}, respectively. The parameters in the input and output embedding layers (yellow background) are not slimmable as dimensionality of the input and output features remains the same.}
	\vspace{-15pt}
	\label{fig:bst_framework}
\end{figure*}

Taking a sequence of input tokens, the original Transformer \cite{vaswani2017attention} adopts an encoder-decoder architecture. The encoder is composed of a cascade of Transformer layers in depth to obtain the {bidirectional} representations by jointly conditioning on both the left and right contexts, and the decoder takes the representations from the last encoder layer as input to guide the learning of {unidirectional} representations by conditioning only on the left context. After that, pure encoder architectures (\emph{e.g.}, BERT \cite{devlin2018bert}) and pure decoder architectures (\emph{e.g.}, GPT \cite{radford2018improving}) are introduced and integrated with the self-supervised pretraining paradigm, which has been used in a wide range of NLP tasks.

\subsection{The BST Framework}\label{sec:the_bst_framework}
Let an $L$-layer Transformer with hidden dimensionality $D$ be the reference model, where $D$ and $L$ denote the width and depth of the model, respectively. The goal of BST is to obtain a single \emph{slimmable} Transformer model that can adaptively adjust to a set of submodels of different widths and depths in the inference stage. In the following, we introduce the width slimming and depth slimming strategies. An overview of the BST framework is shown in Fig. \ref{fig:slim_strategy}.

\noindent\textbf{Width Slimming.} With width slimming, we aim to make each Transformer layer adapt to a set of width slimming ratios with respect to the hidden dimensionality $d$ of the reference model. To achieve this goal, we split the parameters of the reference model into different submodels, with each sharing a specific portion of its model parameters. As shown in Eqs.(\ref{eq:mha}), (\ref{eq:ffn}), and (\ref{eq:ln}), the learnable parameters in the MHA and FFN modules are all linear projections, which can be simply split into submatrices with respect to the given ratios. Inspired by the settings in \cite{hou2020dynabert}, we define the candidate width set as $\mathcal{D}=\{{1}/{4}D, {1}/{2}D, {3}/{4}D, D\}$.

Given a slimmed width $d\in\mathcal{D}$, we need to prune the model parameters in the MHA, FFN, and LN modules according to $d$. In the MHA module, the query, key, and value parameters over $H$ heads can be represented as tiled matrices $W^Q, W^K, W^V\in\mathbb{R}^{D\times D_H*H}$. Given the slimmed width $d$, we keep $D_H$ as a constant and adjust the input dimensionality $D$ and the number of heads $H$ accordingly. This results in the slimmed model parameters $\in\mathbb{R}^{d\times D_H*\hat{H}}$, where $\hat{H}=H*d/D$ refers to the reduced number of heads with the last few heads neglected. The model parameters $W^o$, $W_1$, $W_2$, $\gamma$, and  $\beta$ in the FFN and LN modules are adjusted in a similar manner by slimming the dimensionality of the input and output features to $d$.

In addition to the strategy introduced above, another width slimming strategy was investigated in \cite{hou2020dynabert}. In contrast to our \emph{slim-all} strategy that slims the dimensionality of the input, output, and intermediate representations simultaneously, they introduced an alternative \emph{slim-intermediate} strategy that slims the dimensionality of the intermediate representation while keeping the rest unchanged. The \emph{slim-intermediate} strategy results in a \emph{bottleneck} structure with a low dimensional intermediate representation. This may break the carefully-designed structure of the Transformer (\emph{e.g.}, the $D\rightarrow 4D\rightarrow D$ structure in FFN).

\noindent\textbf{Depth Slimming.} With depth slimming, we aim to make the slimmed submodels adapt to a set of depth slimming ratios with respect to the maximum depth of the reference model. As the depth of a typical Transformer model is usually set to a multiple of 6, we define the candidate depth set as $\mathcal{L}=\{{1}/{6}L, {1}/{3}L, {2}/{3}L, L\}$. Compared with the settings in \cite{hou2020dynabert}, we explore the submodels with much shallow depth, resulting in more compact submodels suitable for mobile devices.  

To perform depth slimming, we first need to determine which layers are to be slimmed given a specific slimming depth $l\in\mathcal{L}$. As shown on the right side of Fig. \ref{fig:slim_strategy}, we first assign an importance score to each layer using certain scoring strategies. After that, we select the top-$l$ layers with the largest scores and keep their original order. Here, we introduce three scoring strategies, which result in three types of slimming strategies: 1) the \emph{slim-random} strategy is the most straightforward strategy, as it simply sets the importance scores to random values; 2) the \emph{slim-first} (or \emph{slim-last}) strategy sets the importance scores in ascending (or descending) order; 3) the \emph{slim-middle} strategy sets the smallest scores to the middlemost layer and gradually increases the scores as it move toward the top and bottom layers. This strategy was inspired by the empirical studies in \cite{wang2020rethinking}, in which the layers closer to the input and output are more important than the middle layers in the Transformer. We use the {slim-middle} strategy as the default option.

\subsection{Integrating BST with Off-the-Shelf VQA Models}
BST is a general framework that can be integrated with arbitrary Transformer-based VQA models in theory. In this paper, we choose three typical Transformer-based models, \emph{i.e.}, MCAN \cite{yu2019mcan}, UNITER \cite{chen2019uniter} and CLIP-ViL\cite{shen2021much}, shown in Fig. \ref{fig:bst_vqa}, to integrate with the proposed BST framework. Without loss of generality, the BST framework can also be applied to other Transformer-based models beyond the VQA task. Due to space limitations, we will not expand the description further.

\noindent\textbf{MCAN$_\texttt{BST}$.} MCAN was the winning solution in the VQA Challenge 2019. It introduces an encoder-decoder-based Transformer architecture to model complex multimodal interactions and perform accurate visual reasoning. Specifically, the input question is encoded as a sequence of word embeddings using pretrained GloVE embeddings followed by an LSTM network. The input image is encoded as a group of object embeddings using a pretrained object detector \cite{anderson2017up-down}. After that, the multimodal embeddings are passed through an $L$-layer encoder-decoder to obtain the attended output features. In the $L$-layer question encoder, the word embeddings are transformed with a self-attention mechanism to obtain the attended question features of the same word length. The attended word features and visual embeddings are further fed into an $L$-layer image decoder to obtain the attended image features with a guided-attention mechanism. On top of the attended question features and image features,  two attentional reduction modules are devised to obtain a question feature and an image feature, respectively. Finally, the two feature vectors are simply fused and then fed to a linear classifier to predict the answer. The MCAN model can be trained from scratch in an end-to-end manner on a specific VQA dataset such as VQA-v2 \cite{goyal2016making}.

Since MCAN's core components are the standard Transformer layers, the model can be seamlessly integrated with the BST framework to obtain a slimmable MCAN$_\texttt{BST}$ model. The width slimming strategy can be directly applied to each encoder and decoder layer in MCAN, and different depth slimming strategies can also be applied to drop a portion of the encoder and decoder layers simultaneously. Furthermore, the model parameters in the attention reduction module on top of the encoder-decoder are derived from two-layer MLPs, which can also be slimmed in width. 

\noindent\textbf{UNITER$_\texttt{BST}$.} UNITER is a representative vision-and-language pretraining (VLP) approach with an $L$-layer Transformer encoder as its backbone. In contrast to MCAN's \emph{training-from-scratch} mechanism, UNITER utilizes a \emph{vision-language pretraining} strategy to learn a generalized backbone model from massive image-text pairs, and then finetunes the backbone to adapt to different multimodal tasks. Specifically, for the VQA task, a task-specific head is appended on top of the backbone so that the representation of the predefined [$\texttt{cls}$] token is fed to a linear classifier to predict the answer. Based on the finetuned UNITER model for VQA, both width slimming and depth slimming are applied to its backbone to transform it into UNITER$_\texttt{BST}$. 

\noindent\textbf{CLIP-ViL$_\texttt{BST}$.} CLIP-ViL shares the same Transformer architecture with UNITER but introduces a more powerful visual encoder to extract visual representations. Specifically, its visual encoder corresponds to a pretrained ResNet-50$\times$4 model, which is pretrained on 400M image-text pairs by CLIP \cite{radford2021learning}. 

Note that the embedding layers in the above models are not slimmable, making the input dimensionality of the first Transformer layer unadjustable. This contradicts our width-slimming strategy. To address this issue, we insert a slimmable linear layer $W_\mathrm{emb}\in\mathbb{R}^{D\times D}$ between the embedder and backbone, and make it slimmable in width to adapt to the BST framework\footnote{For the UNITER$_\texttt{BST}$ and CLIP-ViL$_\texttt{BST}$ models with pretrained model parameters, $W_\mathrm{emb}$ is initialized with an identity matrix and updated with the entire model in an end-to-end manner.}.

\noindent\textbf{Submodel Complexity Analysis.} Given a reference MCAN (UNITER or CLIP-ViL) model of width $D$ and depth $L$, its model size and FLOPs are both approximately proportional to $O(D^2L)$ . This indicates that the computational cost of the smallest submodel $a({1}/{4}D$,${1}/{6}L)$ can be up to 96$\times$ smaller than that of the reference model. In practice, the scaling ratio between the submodels and the reference model is not that large. As shown in Fig. \ref{fig:bst_vqa}, the slimming strategies are only performed in the backbone while the embedders and the classifier are not involved. Their existence introduces an inescapable cost for all the slimmed submodels, limiting the computational overhead of the small submodels. More detailed results are given and analyzed in section \ref{sec:exp}.

\subsection{Training Strategy for BST Models}
The training procedures of BST consists of a \emph{submodel architecture selection} stage and a \emph{knowledge distillation training} stage.

\noindent\textbf{Submodel Architecture Selection.} By combining each width in $\mathcal{D}$ with each depth in $\mathcal{L}$, we obtain a set of submodel architectures $\mathcal{A}$ of different widths and depths as follows:
\begin{equation}\label{eq:comb}
\mathcal{A}=\texttt{combination}(\mathcal{D}, \mathcal{L})
\end{equation}
where $|\mathcal{A}| = |\mathcal{D}|$*$|\mathcal{L}|$. Each architecture $a(d, l)\in \mathcal{A}$ corresponds to a combination of a specific width $d\in\mathcal{D}$ and depth $l\in\mathcal{L}$. In contrast to previous works that maintain all possible submodel architectures \cite{wang2020hat}\cite{hou2020dynabert}, we hypothesize that not every submodel architecture is effective. By the effectiveness of a submodel architecture, we mean the extent to which its computational cost (\emph{e.g.}, in terms of FLOPs or model size) matches its delivered performance after model training. Therefore, devising a heuristic strategy to eliminate such ineffective architectures before BST training can reduce the training costs while improving the performance of the remaining submodels.

According to previous studies on designing efficient Transformers \cite{khetan2020schubert}\cite{hou2020dynabert}, \emph{deep and narrow} architectures usually deliver better performance than \emph{shallow and wide} architectures under constrained computational costs. This principle can be explained in two ways: 1) The Transformer requires a relatively deep model to guarantee good performance; and 2) the computational cost of a Transformer model is proportional to $O(LD^2)$, suggesting that increasing depth is more economical than increasing width.

\begin{figure}
\begin{center}
\includegraphics[width=\columnwidth]{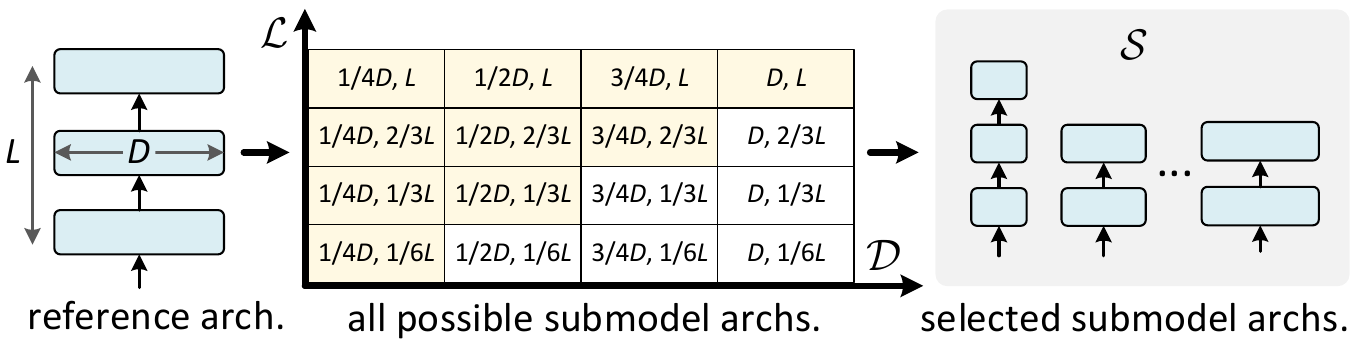}
\vspace{-5pt}
\caption{The diagram of submodel architecture selection by using the triangle selection filtering strategy. The selected submodel architectures (highlighted with yellow background) follow the \emph{deep-and-narrow} principle, which is considered to be more effective than the rest (white background).}
\vspace{-10pt}
\label{fig:triangle}
\end{center}
\end{figure}

To quantize this {deep-and-narrow} principle, we introduce a simple \emph{triangle selection strategy} to filter out the {shallow and wide} submodel architectures. As shown in Fig. \ref{fig:triangle}, we introduce a 2-D indicator matrix $I\in\{0,1\}^{|\mathcal{D}|\times |\mathcal{L}|}$ to track the selection status for all the submodel architectures $\mathcal{A}$. $I(d, l)=1$ indicates that the submodel architecture $a(d, l)$ is selected, and it is 0 otherwise. The indicator matrix $I$ is first initialized with all-one values and then converted to an \emph{upper-triangle} matrix. This strategy allows us to obtain \emph{six} {shallow and wide} submodel architectures, which correspond to the matrix elements above the main diagonal. The submodel architectures $\mathcal{S}$ selected from $\mathcal{A}$ are defined as follows:
\begin{equation}\label{eq:submodels}
{\mathcal{S}}=\texttt{upper-triangle}(\mathcal{A})=\{a(d, l) |I(d,l)=1 \}
\end{equation}

\noindent\textbf{Knowledge Distillation Training.} After obtaining the selected submodel architectures ${\mathcal{S}}$, we train a slimmable model $M_\texttt{BST}$ that can elastically adapt to any submodel architecture $a\in\mathcal{S}$, where $M$ refers to any of the above VQA models. Given the BST model $M_\texttt{BST}$ and a submodel architecture $a$, the obtained submodel is denoted as $M_\texttt{BST}^{(a)}$.

To obtain the BST model, we introduce a knowledge-distillation (KD) training mechanism as follows. In general, we first train an ordinary model ${M_\texttt{teacher}}$ as the teacher model without model slimming and freeze its model parameters. After that, the BST model $M_\texttt{BST}$ is initialized with the model parameters from ${M_\texttt{teacher}}$, and it can be viewed as the student model. Each slimmed submodel $M_\texttt{BST}^{(a)}$ shares a specific portion of the model parameters from $M_\texttt{BST}$ and is trained with the supervision of the teacher model using the KD strategy \cite{hinton2015distilling}. Specifically, given a training sample, we feed it through the teacher model and each slimmed submodel simultaneously. The predicted answer distribution from the teacher model is regarded as a soft label, and a proper loss function is imposed on the soft label and each submodel prediction to train the corresponding model\footnote{Different loss functions (\emph{e.g.}, BCE and KL-divergence) can be used as the KD loss flexibly, depending on the loss function used in the teacher model.}.

Unlike the existing approaches \cite{hou2020dynabert} that decouple the width slimming and depth slimming into two training stages, we use a simpler one-stage training paradigm for BST with standard mini-batch SGD. In each iteration, we select $K$ submodel architectures from $\mathcal{S}$ and feed the same input samples to both the teacher model and the selected $K$ submodels to obtain ($K+1$) predictions in total. After that, we apply the KD loss between the predictions of the teacher and each of the $K$ student models and use the accumulated back-propagated gradients to update the model parameters of $K$. To stabilize the BST training, the $K$ selected submodels consist of two determined submodels (the smallest one and the largest one) and another ($K$-2) randomly sampled submodels. We use $K$=4 as the default setting in our experiments.

The whole BST training process is shown in Algorithm \ref{alg:1}.

\begin{algorithm}
\small
\SetAlgoLined
\KwIn{A reference Transformer architecture with width $D$ and depth $L$. Two predefined sets $\mathcal{D}$ and $\mathcal{L}$ define the slimming widths and depths w.r.t. $D$ and $L$, resp. The number of sampled submodel architectures $K$ per training iteration. $a_s$ and $a_l$ refer to the smallest and largest submodel architectures, respectively.}
\KwOut{An optimized BST model $M_\mathrm{BST}$.}
\emph{Stage I: Submodel Architectures Selection}\;
$\mathcal{A} = \texttt{combination}(\mathcal{D},\mathcal{L})$\;
${\mathcal{S}}$ = $\texttt{upper-triangle}(\mathcal{A})$\;
\emph{Stage II: Knowledge Distillation Training}\;
Train $M_\texttt{teacher}$ to obtain its optimized model parameters $\theta$\;
Initialize the model parameters $\theta_\texttt{BST}$ for $M_\texttt{BST}$ with $\theta$\;
\For{$i=1$ \KwTo {max-iter}}{
        Randomly sample a mini-batch of data $x$\;
        Initialize the sampled architecture set $\Omega=\emptyset$\;
        \# add the smallest and largest submodel architectures.\\
        $\Omega\leftarrow\Omega \cup \{a_{s}, a_{l}\}$\;
        \# add another $K-2$ architectures via random sampling.\\
        \For{$j=1$ \KwTo $K-2$}{
        Randomly sample a submodel architecture $a\sim\mathcal{S}_{\backslash{\Omega}}$ \;
        $\Omega\leftarrow\Omega \cup a$
        }
        \# submodel training using knowledge distillation.\\
        Feed-forward the teacher model: $y=M_\texttt{teacher}(x)$\;
        Freeze $M_\texttt{teacher}$ by stopping gradients: $y.\texttt{detach}()$\;
        \ForEach{$a\in\Omega$}{
         Feed-forward the submodel: $\hat{y}=M_\texttt{BST}^{(a)}(x)$\;
         Compute loss: $loss=\texttt{KD}(y,\hat{y})$\;
         Accumulate backward gradients: $loss.\texttt{backward}()$\;
        }
        Update model parameters $\theta_\texttt{BST}$.
}
 \caption{Training procedure for BST.}\label{alg:1}
\end{algorithm}

\subsection{In-Depth Comparison of BST and DynaBERT}
As mentioned above, our BST framework has close connections with DynaBERT \cite{hou2020dynabert}. We conduct an in-depth comparison and describe their differences in terms of methodology.

In terms of the width-slimming strategy, DynaBERT adopts a \emph{slim-intermediate} strategy that only reduces the dimensionality of the intermediate representation while keeping the input and output representations unchanged, which may break the carefully designed bottleneck structure in the original Transformer architecture. In contrast, we use a \emph{slim-all} strategy that reduces all the dimensionalities, keeping the bottleneck structure to achieve better performance.
 
In terms of the depth-slimming strategy, DynaBERT uses a simple slimming strategy by slimming the layers uniformly. In contrast, BST introduces a \emph{slim-middle} strategy that considers the layer importance and prefers to slim the middlemost layers first, which facilitates model performance. 

In terms of the training strategy, DynaBERT uses a two-stage training paradigm in which the width and depth slimming are learned separately, with all submodels being updated in each training step. In contrast, BST uses a simpler one-stage training paradigm to learn width and depth slimming simultaneously and sample a small number of submodels in each step, which significantly improves the training efficiency. Moreover, DynaBERT maintains the submodel architectures of all the width-height combination during training which may include redundant ones. In contrast, BST additionally introduces a submodel selection strategy to remove ineffective submodel architectures before training. This strategy not only reduces the training costs but also improves the performance of the remaining submodels

To summarize, BST has advantages over DynaBERT in terms of training efficiency and model performance. More quantitative comparisons are provided in section \ref{sec:exp}.

\section{Experimental Results}\label{sec:exp}
In this section, we present experiments to evaluate the performance of our BST framework on two benchmark VQA datasets, namely, VQA-v2 \cite{antol2015vqa} and GQA \cite{hudson2019gqa}. As mentioned above, we integrate BST with three typical Transformer-based VQA models, namely, MCAN \cite{yu2019mcan}, UNITER \cite{chen2019uniter}, and CLIP-ViL \cite{shen2021much}, to demonstrate the effectiveness and generality of BST. Furthermore, we conduct extensive ablation experiments on VQA-v2 to explore the effects of different components.

\begin{table*}
\normalsize
        \centering
        \caption{Comparison to the state-of-the-art approaches on VQA-v2. For a fair comparison, the compared methods are split into two groups depending on whether they are trained from scratch or pretrained on external data (separated by a double-line). The number of parameters is calculated from an entire model, including the backbone, input and output embedding layers. The number of FLOPs is calculated from one single sample. $^\dag$: the UNITER$_\texttt{DYN}$ model refers to another slimmable model that integrates UNITER and DynaBERT \cite{hou2020dynabert}, which is reimplemented by ourselves.}\label{tab:sota_vqav2}
        \begin{tabular}{c|l|rr|cccc|cccc}
        \toprule
      \multirow{3}{*}{\#}   & \multirow{3}{*}{model}& \multirow{3}{*}{\#params} & \multirow{3}{*}{\#FLOPs} & \multicolumn{4}{c|}{\texttt{test-dev}} &\multicolumn{4}{c}{\texttt{test-std}} \\
\cmidrule{5-12}
& &&& All & Y/N & Num & Other & All &Y/N & Num & Other \\
\midrule
 \multicolumn{12}{l}{\emph{From-scratch training with augmented Visual Genome data}}\\
\midrule
1 & UpDn \cite{anderson2017up-down} & 22M & 1.1G & 65.32& 81.82& 44.21& 56.05 &65.67 & 82.20 & 43.90 & 56.26\\
2&MFB \cite{yu2017mfb} & 68M & 2.4G & 68.40 & 84.78 & 49.05 & 58.82 & - & - & - & - \\
3 &MFH \cite{yu2018beyond} & 102M & 2.5G & 68.76 & 84.27 &49.56 &59.89 & - & - & - & -\\
4 & BAN \cite{kim2018bilinear} &112M&12.3G&69.66& 85.46 &50.66& 60.50&-&-&-&-\\
5&MUAN \cite{yu2019multimodal}& 83M & 17.3G & 70.82 &86.77& \textbf{54.40}& 60.89& 71.10& - & - & -  \\
6& MCAN($D$=512,$L$=6) \cite{yu2019mcan} & 58M & 5.5G & 70.63 & 86.82 & {53.26} & 60.72  & 70.90 & - & - & - \\
7& MCAN(${1}/{4}D,{1}/{3}L)$) \cite{yu2019mcan} & 10M & 0.2G & 67.19&	83.38&	49.75&	57.31&-&-&-&-\\
8 & RWSAN \cite{qin2022deep} & 20M& 6.5G & 70.19& 86.45& 52.18& 60.38 & -&-&-&- \\
\midrule
9 & {MCAN$_\texttt{BST}(D,L)$} & 58M & 5.5G & \textbf{71.05} & \textbf{87.39} & 52.96 & \textbf{61.19} & \textbf{71.28} & \textbf{87.36} & \textbf{52.77} & \textbf{61.52}\\
10& {MCAN$_\texttt{BST}({1}/{2}D,L)$}& 22M & 1.5G & 70.45 & 86.84 & 52.89 & 60.43 & - & - & - & - \\
11& {MCAN$_\texttt{BST}({1}/{2}D,{1}/{3}L)$}& 14M  & 0.6G & 69.42 & 85.68 &51.96 & 59.48 & - & - & - & -  \\
12& {MCAN$_\texttt{BST}({1}/{4}D,{1}/{3}L)$}& 10M & 0.2G & 68.16 & 84.84 & 50.27 & 57.95 & - & - & - & - \\
\midrule
\midrule
\multicolumn{12}{l}{\emph{Vision-language pretraining with massive external data}}\\
 \midrule
13 & ViLBERT \cite{lu2019vilbert} & 221M & 18.7G & 70.55  & - & - & -  & 70.92 & - & - & - \\
14 & VLBERT \cite{su2019vl}      & 116M & 20.7G & 71.16  & - & - & -  & - & - & - & - \\
15 & LXMERT \cite{tan2019lxmert} & 183M & 20.3G & 72.42 & - & - & - & 72.54 & 88.20 & 54.20 & 63.10\\
16 & OSCAR  \cite{li2020oscar}           & 116M & 38.6G & 73.16 & - & - & - & 73.44 & - & - & -\\
17 & ALBEF \cite{li2021align} & 210M & 77.9G & 75.84 & - & - & - & 76.05 & \textbf{91.67} & 55.43 & 67.19\\
18 & UNITER($D$=768,$L$=12)  \cite{chen2019uniter}               & 117M & 20.2G & 72.70& 88.86& 55.10 &62.87& 72.95 &89.00& 55.37 &63.01\\
19 & UNITER(${1}/{4}D,{1}/{3}L)$)  \cite{chen2019uniter}               & 33M&	0.8G&	66.89	&83.25&	49.97&	56.74&-&-&-&-\\
20 & CLIP-ViL($D$=768,$L$=12) \cite{shen2021much} & 237M & 82.1G & \textbf{76.44} &  91.38 & 58.12 & \textbf{67.86} & \textbf{76.74} & {91.60} & 58.09 & \textbf{68.07}\\
21 & CLIP-ViL(${1}/{4}D,{1}/{3}L)$) \cite{shen2021much} & 153M&	59.3G&	70.32&	85.40&	52.12&	61.59 &-&-&-&-\\
\midrule
22 & $^\dag$UNITER$_\texttt{DYN}(D,L)$ & 117M  &20.2G & 73.19    & 89.02& 56.39& 63.46 &-&-&-&-\\
23 & $^\dag$UNITER$_\texttt{DYN}(3/4D, 1/2L)$ & 64M  & 8.1G & 71.99 &87.87&54.76 & 62.34&-&-&-&-\\
24 & $^\dag$UNITER$_\texttt{DYN}(1/4D, 1/2L)$ & 43M    &3.1G & 70.48  & 86.37 &52.72 & 60.93&-&-&-&-\\
\midrule
25& {UNITER$_\texttt{BST}(D,L)$} & 117M & 20.2G & {73.27} & {89.01} & {56.73} & {63.57} & {73.46} & {89.17} & {56.28} & {63.73}\\
26 & {UNITER$_\texttt{BST}({1}/{2}D,L)$} & 53M & 5.6G & 72.11 & 87.83 &55.59 & 62.42 & - & - & - & -\\
27 & {UNITER$_\texttt{BST}({1}/{2}D,{1}/{3}L)$} & 39M & 2.2G & 70.65 & 86.47 &53.47  &61.02 & - & - & - & -\\
28 & {UNITER$_\texttt{BST}({1}/{4}D,{1}/{3}L)$} & 33M & 0.8G & 69.68 &85.49 & 52.26 &  60.12 & - & - & - & -\\
\midrule
29 & CLIP-VIL$_\texttt{BST}(D,L)$ & 237M&82.1G&	\textbf{76.44}&	\textbf{91.44}&	\textbf{58.27}&	{67.78}&	{76.70}&	{91.54}&	\textbf{58.71}&	{67.90}\\
30 & CLIP-VIL$_\texttt{BST}({1}/{2}D,L)$ & 172M	&64.9G&	75.17&	90.29&	57.31&	66.30&-&-&-&-\\
31 & CLIP-VIL$_\texttt{BST}({1}/{2}D,1/3L)$ & 158M&	60.8G&	73.86&	89.20&	55.43&	64.95&-&-&-&-\\
32 & CLIP-VIL$_\texttt{BST}({1}/{4}D,1/3L)$ & 153M&	59.3G&	72.34&	87.41	&54.11&	63.63&-&-&-&-	\\
\bottomrule
 \end{tabular}
\end{table*}
\subsection{Datasets}
\textbf{VQA-v2} is the most commonly used VQA dataset \cite{goyal2016making}. It contains human-annotated QA pairs for MS-COCO images \cite{lin2014microsoft}. The dataset is split into three sets: \texttt{train} (80k images with 444k questions); \texttt{val} (40k images with 214k questions); and \texttt{test} (80k images with 448k questions). The \texttt{test} set is further split into \texttt{test-dev} and \texttt{test-std} sets. The reported results include three per-type accuracies (yes/no, number, and other), as well as an overall accuracy.

To make a fair comparison among the compared models, we follow the dataset splitting strategy in UNITER that further splits the \texttt{val} set into a \texttt{minival} subset of 5k images and a \texttt{trainval} subset of the remaining 35k images \cite{chen2019uniter}. All the results reported in the experiments come from the models that are trained on the augmented \texttt{train+trainval+vg} sets, where \texttt{vg} denotes the augmented VQA samples from Visual Genome \cite{krishna2016visual}. The obtained models are validated on the \texttt{minival} set offline, and evaluated on the \texttt{test-dev} and \texttt{test-std} sets online.

\textbf{GQA} is a challenging VQA dataset that requires more complex reasoning skills \cite{hudson2019gqa}. It consists of 113K images and 1.2M balanced question-answer pairs of assorted types and varying compositionality degrees, measuring performance on an array of reasoning skills such as object and attribute recognition, spatial reasoning, logical inference, and comparisons. The dataset is split into the following four sets: \texttt{train} (72k images with 943k questions), \texttt{val} (10k images with 132k questions), \texttt{test-dev} (398 images with 12k questions), and undisclosed \texttt{test-challenge} (1.6k images with 50k questions). Following the suggestions in the official GQA guideline\footnote{\url{https://cs.stanford.edu/people/dorarad/gqa/evaluate.html}}, all the models are trained on the \texttt{train+val} sets and evaluated on the \texttt{test-dev} set.

\subsection{Experimental Setup}
The model architectures and training hyperparameters of the MCAN$_\texttt{BST}$, UNITER$_\texttt{BST}$, and CLIP-ViL$_\texttt{BST}$ models are the same as those in their original models  \cite{yu2019mcan}\cite{chen2019uniter}\cite{shen2021much}, respectively. 

For MCAN$_\texttt{BST}$, the hidden dimensionality $D$, number of heads $H$, and number of layers $L$ are set to 512, 8, and 6, respectively. The MCAN$_\texttt{BST}$ models are trained on the VQA-v2 and GQA datasets using slightly different settings. On VQA-v2, binary cross-entropy (BCE) is used as the loss function for both the teacher and the BST model, and both models are trained for up to 15 epochs with a batch size of 64 and a base learning rate of 1e-4.  The learning rate is warmed-up for 3 epochs and decays by 1/5 every 2 epochs after 10 epochs. On GQA, the learning rate and batch size are the same as those on VQA-v2. KL-divergence is used as the loss function and the teacher and BST models are trained for 11 epochs. The learning rate is warmed-up for 2 epochs and decays by 1/5 every 2 epochs after 8 epochs.

For UNITER$_\texttt{BST}$ and CLIP-ViL$_\texttt{BST}$, we adopt the network architecture from the BERT-base model with $D=768$, $H=12$, and $L=12$. Given a model the finetuned on VQA-v2 as the teacher, UNITER$_\texttt{BST}$ and CLIP-ViL$_\texttt{BST}$ are initialized from their corresponding teacher models and trained using the AdamW optimizer. UNITER$_\texttt{BST}$ is trained for up to 130k iterations with a batch size of 5,120 and a base learning rate of 1.5e-4. CLIP-ViL$_\texttt{BST}$ is trained for up to 15 epochs with a batch size of 32 and a base learning rate of 2e-4. To reduce the usage of GPU memory, we freeze the model parameters in the visual encoder during BST training.

\begin{table}
	\normalsize
	\centering
	\caption{Accuracies of the state-of-the-art methods on GQA. All entries use the officially provided object features for images and are evaluated on the \texttt{test-dev} split.}\label{tab:sota_gqa}
	\begin{tabular}{l|rr|c}
		\toprule
		model     & \#params & \#FLOPs  &  accuracy \\
		\midrule
		UpDn \cite{anderson2017up-down} & 30M & 2.8G & 51.62   \\
		BAN \cite{kim2018bilinear} & 120M & 14.9G & 55.81   \\
		MCAN($D$=512,$L$=6) \cite{yu2019mcan} & 59M & 6.5G & 56.64   \\
		\midrule
		{MCAN$_\texttt{BST}(D,L)$} & 59M & 6.5G & \textbf{57.83}  \\
		{MCAN$_\texttt{BST}({1}/{2}D,L)$} & 22M & 1.9G & 57.67  \\
		{MCAN$_\texttt{BST}({1}/{2}D,{1}/{3}L)$} & 15M & 0.8G & {57.09}  \\
		{MCAN$_\texttt{BST}({1}/{4}D,{1}/{3}L)$} & 10M & 0.4G & 56.38  \\
		\bottomrule
	\end{tabular}
\end{table}

\subsection{Main Results}
In Tables \ref{tab:sota_vqav2} and \ref{tab:sota_gqa}, we compare MCAN$_\texttt{BST}$, UNITER$_\texttt{BST}$, and CLIP-ViL$_\texttt{BST}$ to the state-of-the-art VQA methods on VQA-v2 and GQA, respectively. For MCAN$_\texttt{BST}$, the compared methods include UpDn \cite{anderson2017up-down}, MFB \cite{yu2017mfb}, MFH \cite{yu2018beyond}, BAN \cite{kim2018bilinear}, MUAN \cite{yu2019multimodal}, and MCAN \cite{yu2019mcan}, which were the best-performing solutions in the VQA Challenge in recent years. In addition, we introduce the lightweight VQA model RWSAN \cite{qin2022deep} into the comparison. For UNITER$_\texttt{BST}$ and CLIP-ViL$_\texttt{BST}$, the compared methods include ViLBERT \cite{lu2019vilbert}, VLBERT \cite{su2019vl}, LXMERT \cite{tan2019lxmert}, OSCAR \cite{li2020oscar}, ALBEF \cite{li2021align}, UNITER \cite{chen2019uniter} and CLIP-ViL \cite{shen2021much}, which are representative vision-language pretraining methods. Moreover, although DynaBERT \cite{hou2020dynabert} is not specifically designed for VQA, we reimplement it ourselves to integrate it with UNITER, resulting in a slimmable VQA model UNITER$_\texttt{DYN}$ for comparison.
Due to space limitations, we do not show the results of all the submodels (\emph{i.e.}, the ten submodels selected by the triangle selection strategy in Fig. \ref{fig:triangle}) in these tables. Instead, four typical submodels are selected for comparison with the state-of-the-art approaches. 

\begin{figure}
	\begin{center}
		\includegraphics[width=0.98\columnwidth]{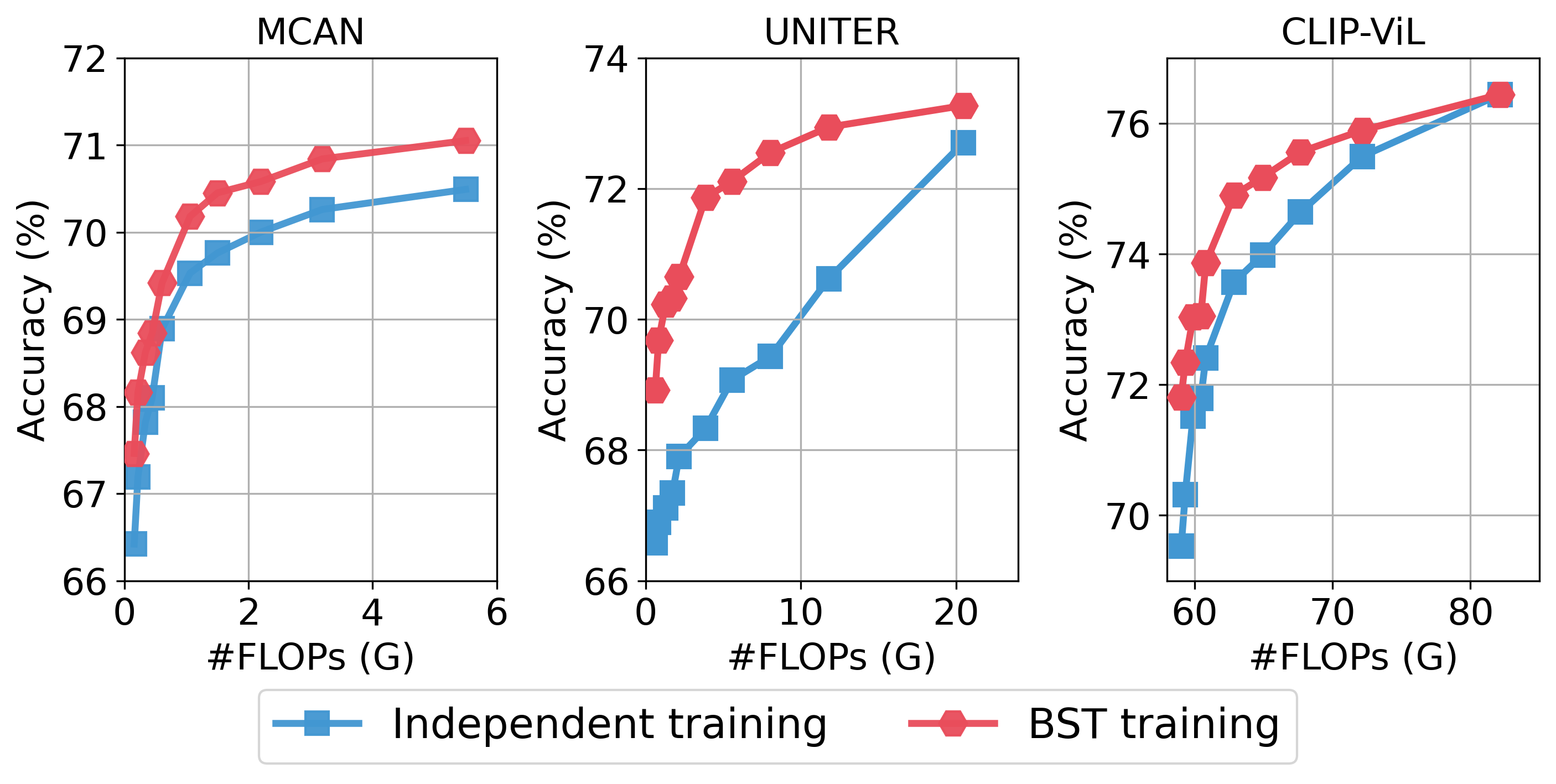}
		\caption{Accuracy \emph{vs.} \#FLOPs on the VQA-v2 \texttt{test-dev} split. For each VQA model (\emph{i.e.}, MCAN \cite{yu2019mcan}, UNITER \cite{chen2019uniter}, and CLIP-ViL \cite{shen2021much}), we report the results of ten submodels obtained from BST training and independent training, respectively.}
		\vspace{-10pt}
		\label{fig:bst_ind_compare}
	\end{center}
\end{figure}

\begin{figure}
	\begin{center}
		\includegraphics[width=0.98\columnwidth]{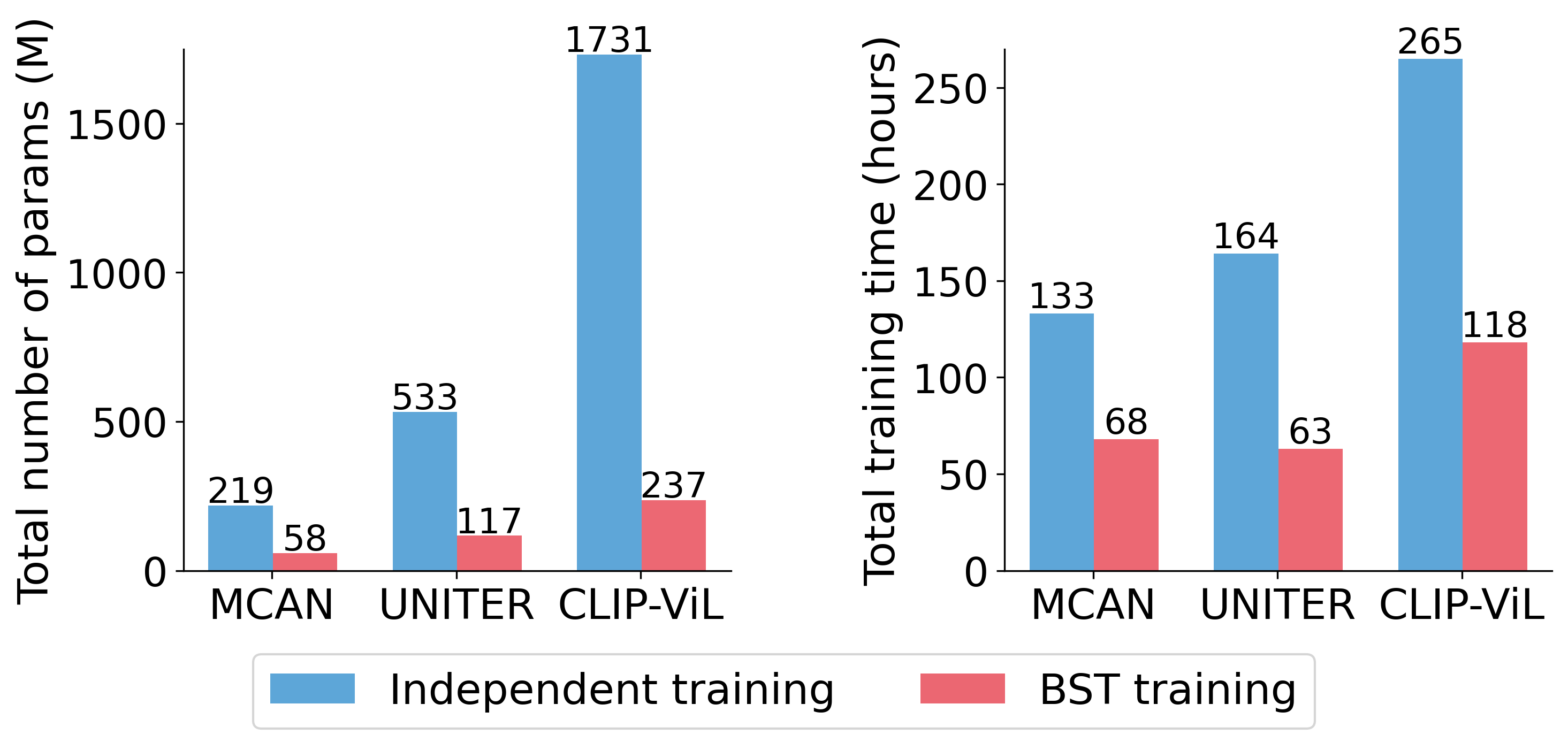}
		\caption{Total number of parameters (left) and total training time (right) of the 10 submodels obtained by BST training and independent training, respectively. The training time is measured on a single Nvidia TitanV GPU.}
		\vspace{-10pt}
		\label{fig:bst_ind_compare2}
	\end{center}
\end{figure}

\begin{table*}
	\centering
	\caption{Runtime latency (ms) of three typical MCAN$_\mathtt{BST}$ submodels on three platforms, namely GPU, CPU, and mobile. For each platform, we report the results on three device.}
	\begin{tabular}{c|l|ccc|ccc|ccc}
		\toprule
		\multicolumn{2}{c|}{platforms}  & \multicolumn{3}{c|}{GPU} & \multicolumn{3}{c|}{CPU} & \multicolumn{3}{c}{mobile}\\
		\midrule
		\multicolumn{2}{c|}{device type} & \makecell{RTX\\3090} & \makecell{RTX\\2080Ti}& \makecell{GTX\\1650}& \makecell{Intel\\E5-2620v4} & \makecell{Intel\\I3-10100} & \makecell{AMD\\R7 4800H} & \makecell{Qualcomm\\Snapdragon 888} & \makecell{MediaTek\\Dimensity 1100} & \makecell{Qualcomm\\Snapdragon 660} \\
		\midrule
		1& $(D,L)$ & 22&29&40&51&101&127&167&361&439\\
		2& $(1/4D,L)$ & 21	&29&	35&	30&	44&	50&	87&	127&197\\
		3& $(1/4D,1/6L)$ & 5&	7&	12&	11&	14&	22&	58&	93&	160\\
		\bottomrule
	\end{tabular}
	\label{table:run_platform}
\end{table*}

From the results in the upper part of Table \ref{tab:sota_vqav2}, we have the following observations: 1) With the same model architecture, the largest submodel MCAN$_\texttt{BST}(D,L)$ and the smallest submodel MCAN$_\texttt{BST}({1}/{4}D,{1}/{3}L)$ outperform their independently-trained counterparts, respectively (\#9 \emph{vs}. \#6, \#12 \emph{vs}. \#7). This improvement is a benefit of the synergistic effect of the weight-sharing submodel architectures and KD training strategy. 2) With only 0.38$\times$ the model size and 0.27$\times$ the FLOPs, MCAN$_\texttt{BST}({1}/{2}D,L)$ is still competitive with the reference MCAN model (\#10 \emph{vs}. \#6), showing the potential of width slimming. In contrast to RWSAN \cite{qin2022deep}, which has a similar model size, MCAN$_\texttt{BST}({1}/{2}D,L)$ achieves higher accuracy and 0.25$\times$ the FLOPs. 3) By slimming the depth to ${1}/{3}L$ (\#11), its corresponding model size and FLOPs are respectively reduced to 0.6$\times$ and 0.4$\times$ those of its counterpart in \#10, respectively, at the expense of a 1-point accuracy drop. Compared with MFB \cite{yu2017mfb}, MFH \cite{yu2018beyond}, and BAN \cite{kim2018bilinear}, MCAN$_\texttt{BST}({1}/{2}D,{1}/{3}L)$ achieves superior or comparable performance with up to 0.125$\times$ the model size and 0.05$\times$ the FLOPs. 4) MCAN$_\texttt{BST}({1}/{4}D,{1}/{3}L)$ still outperforms UpDn \cite{anderson2017up-down} by 2.8 points with an extremely small model size of 10M. This model size is close to the lower bound of MCAN, which includes 7.8M uncompressible model parameters in the input and output embedding layers.

From the results in the lower part of Table \ref{tab:sota_vqav2}, we obtain similar observations to those on MCAN$_\texttt{BST}$. The slimmable UNITER$_\texttt{BST}$ and CLIP-ViL$_\texttt{BST}$ models attain superior or comparable performance to their reference independently-trained counterparts (\#25 \emph{vs.} \#18, \#28 \emph{vs.} \#19, \#29 \emph{vs.} \#20, \#32 \emph{vs.} \#21). The slimmed submodels in \#26-28 and \#30-32 attain significant reduction in computational costs at the expense of a drop in accuracy. Compared with the slimmable model UNITER$_\texttt{DYB}$ of DynaBERT \cite{hou2020dynabert} (\#22-24), our UNITER$_\texttt{BST}$ achieves higher performance with similar model sizes (\#22 \emph{vs.} \#25, \#23 \emph{vs.} \#26, \#24 \emph{vs.} \#27). Additionally, the total training time for DynaBERT is 3.6$\times$ longer than ours. These results verify that our slimming and training strategies are more effective than those of DynaBERT.

To further examine the generalization of BST, we compare MCAN$_\texttt{BST}$ to the state-of-the-art methods on GQA. Table \ref{tab:sota_gqa} shows that MCAN$_\texttt{BST}(D,L)$ is 1.2 points higher than the reference MCAN model without BST. Furthermore, with 0.17$\times$ the model size and 0.06$\times$ the FLOPs, MCAN$_\texttt{BST}({1}/{4}D,{1}/{3}L)$ achieves comparable results to the reference MCAN model, surpassing the rest of its counterparts by a distinct margin.

Next, we show the results of all the ten submodels in terms of BST training and standard independent training. By independent training, we mean that each submodel is trained independently \emph{without} sharing its model parameters\footnote{For the independent training for UNITER and CLIP-ViL, each submodel is first initialized with a specific portion of the model parameters from the pretrained model and then finetuned independently.}. From the results in Fig. \ref{fig:bst_ind_compare}, we can see that all the ten submodels obtained by BST training deliver better performance than their counterparts obtained by independent training. This corroborates the results in Table \ref{tab:sota_vqav2}.

Finally, the submodels obtained by BST are slimmed from \emph{one single model without retraining}, outperforming the same submodels obtained by independent training in terms of both the total model size and total training time. From the results in Fig. \ref{fig:bst_ind_compare2}, we can see that the total model size of the ten submodels obtained by BST training is $\sim$25\% of that obtained by independent training. Furthermore, the total training time for BST training is $\sim$50\% of that for independent training, revealing the synergistic effect of different weight-sharing submodels in BST training.

To summarize, these observations above verify the effectiveness and generality of the proposed BST in terms of different Transformer architectures, training paradigms, and visual encoders.

\subsection{Runtime Latency on Different Hardware Platforms}
To precisely measure the efficiency of different submodels, we report the runtime latency on different hardware platforms in Table \ref{table:run_platform}. Specifically, we deploy three typical submodels of a MCAN$_\texttt{BST}$ model (\emph{i.e.}, the largest submodel, the smallest submodel, and a deep and narrow submodel) on three commonly-used platforms (GPU, CPU, and mobile devices). For each platform, we choose three devices with different computational capabilities.

From the results in Table \ref{table:run_platform}, we can see that: 1) For each submodel, different platforms and device types lead to significant discrepancies in terms of runtime latency, which stems from their diverse computational capabilities. 2) When width slimming is performed (\#1 \emph{vs.} \#2), the latency on the GPU platform is not reduced significantly, while the latency on the CPU platform decreases prominently. This suggests that the GPU has no significant advantage over the CPU for these \emph{narrow} models. 3) When depth slimming is also further performed (\#1 \emph{vs.} \#3), the latencies on the GPU and CPU platforms are not distinct. This suggests that computational capabilities of the GPU and CPU platforms are excessive for such small submodels. 4) The smallest submodel in \#3 can be deployed on a non-latest cellphone with a Snapdragon 888 chip. The 58 ms latency can support applications with real-time requirements.

\captionsetup[subtable]{font=small}
\begin{table*}
	\normalsize
	\caption{Ablations of the MCAN$_\texttt{BST}$ variants models evaluated on the \texttt{minival} split of VQA-v2. The default setting and the best result are bolded.}
	\begin{subtable}[t]{.53\textwidth}
		\centering
	\begin{tabular}{l|c|cc}
		\toprule
		slimming strategy   &submodel &  \#FLOPs & accuracy (\%)\\
		\midrule
		\textbf{slim-all}   &($1/2D$,$L$)&1.5G&\textbf{67.98}\\
		slim-interm. \cite{hou2020dynabert} &($1/4D$,$L$)&1.6G& 67.86\\
		\midrule
		\textbf{slim-all}   &($1/2D$,$1/3L$)&0.6G&\textbf{66.95}\\
		slim-interm. \cite{hou2020dynabert} &($1/4D$,$1/3L$)&0.8G& 66.83\\
		\bottomrule
	\end{tabular}
	\vspace{5pt}
	\subcaption{\textbf{Width Slimming.} Under the same model depth and similar number of FLOPs, the obtained submodels trained with the \emph{slim-all} strategy outperform the \emph{slim-intermediate} strategy, showing the significance of keeping the ratio of input-output dimensionality in width slimming.}
	\label{table:aba_width}
\end{subtable}
\quad
\begin{subtable}[t]{.45\textwidth}
	\centering
	\begin{tabular}{l|c}
		\toprule
		slimming strategy ~~~~~~   & ~~average accuracy (\%)~~ \\
		\midrule
		slim-random & 66.98 \\[0.5mm]
		slim-first & 66.99 \\[0.5mm]
		slim-last & 67.07 \\[0.5mm]
		\textbf{slim-middle} & \textbf{67.14} \\[0.3mm]
		\bottomrule
	\end{tabular}
	\vspace{5pt}
	\subcaption{\textbf{Depth Slimming.} The \emph{slim-middle} strategy outperforms all the counterparts in terms of average accuracy, verifying that the middle layers in Transformer are less important than the first and last ones. More evidence is shown in Fig. \ref{fig:layer_score}.}
	\label{table:aba_depth}
\end{subtable}
\\
\begin{subtable}[t]{.53\textwidth}
	\centering
	\begin{tabular}{l|cc}
		\toprule
		\#sampled submodels  & average accuracy (\%) & training time (h)\\
		\midrule
		\textit{K}=3 & 66.88 & 64\\
		\textbf{\textit{K}=4} & {67.14} & 68\\
		\textit{K}=5 & {67.15} & 73\\
		\textit{K}=6 & \textbf{67.16} & 79\\
		\bottomrule
	\end{tabular}
	\vspace{5pt}
	\subcaption{\textbf{Number of Sampled Submodels.} $K$=4 is a good trade-off between the average accuracy and training time. Further increasing $K$ does not bring prominent performance improvement but takes more training time.}
	\label{table:aba_numsamples}
\end{subtable}
\quad
\begin{subtable}[t]{.45\textwidth}
	\centering
	\begin{tabular}{l|c}
		\toprule
		training strategy    & average accuracy (\%) \\
		\midrule
		w/ random init, w/o KD & 65.89 \\
		w/ random init, w/ ID \cite{yu2019universally} ~& 65.55 \\
		w/ random init, w/ KD & 66.95 \\
		\textbf{w/ teacher init, w/ KD} & \textbf{67.14} \\
		\bottomrule
	\end{tabular}
	\vspace{5pt}
	\subcaption{\textbf{Training Strategy.} The teacher initialization and KD strategies show advantages over the random initialization and ID strategies in terms of average accuracy, respectively.}
	\label{table:aba_train}
\end{subtable}
\vspace{-5pt}
\label{table:aba}
\end{table*}

\subsection{Ablation Studies}\label{sec:aba}
\begin{figure}
	\begin{center}
		\begin{overpic}[width=0.49\textwidth]
			{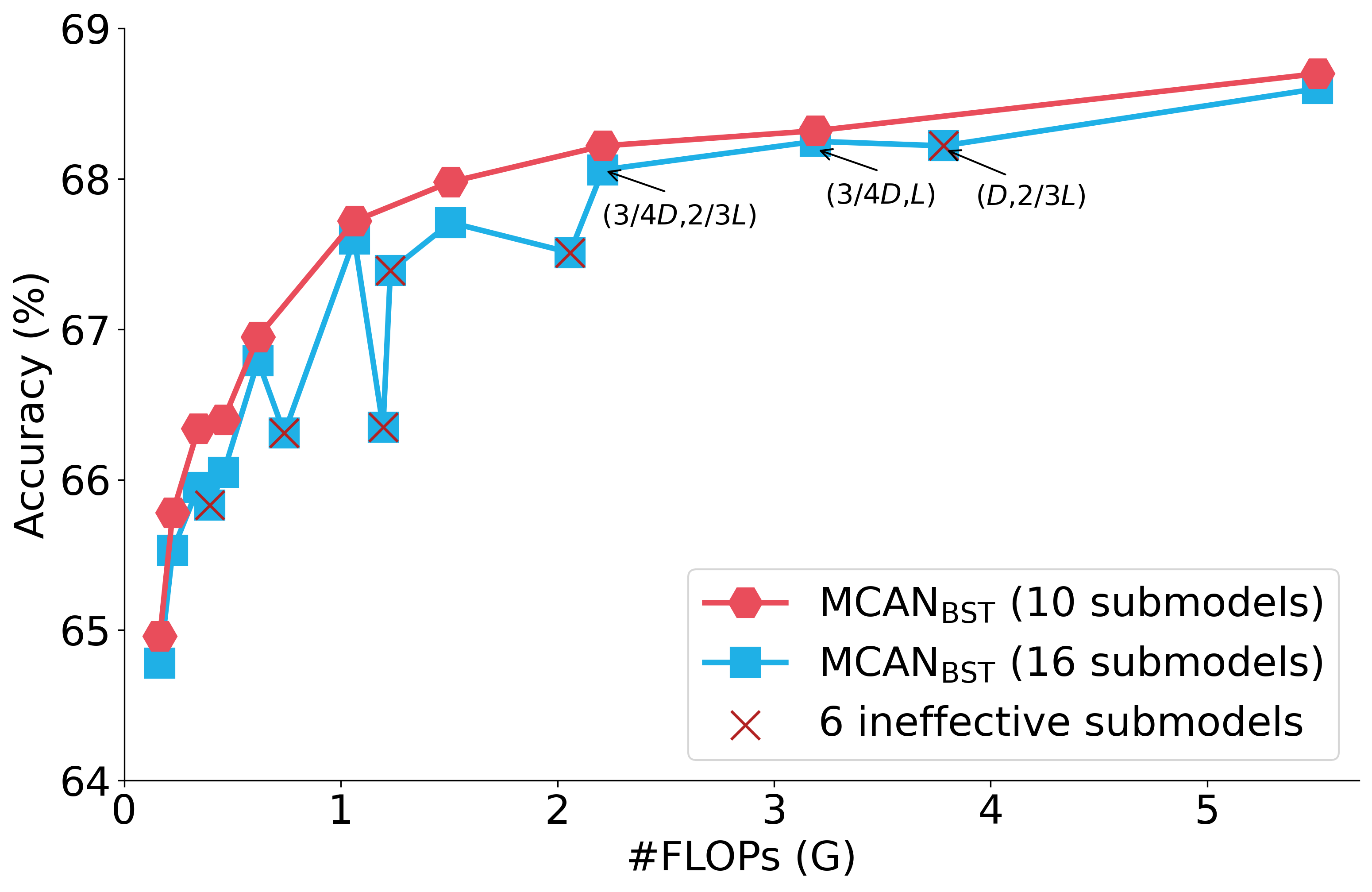}
			\put(45,23){\includegraphics[width=0.27\textwidth]{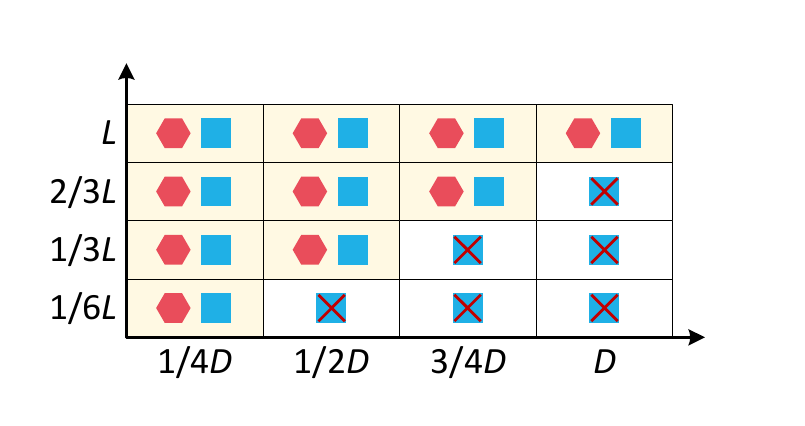}}
		\end{overpic}
		\caption{\textbf{Submodel Selection.} The accuracies \textit{vs}. FLOPs on VQA-v2 \texttt{minival} split are reported to compare two MCAN$_\texttt{BST}$ variants with 10 and 16 submodels, respectively. The network architectures of the 10 submodels (red squares) correspond to a subset of the 16 submodels (blue squares) after discarding 6 \emph{ineffective} submodels (red crosses) by using our triangle selection strategy.}\label{fig:aba_selection}
	\end{center}
	\vspace{-15pt}
\end{figure}

We run a number of ablations on MCAN$_\texttt{BST}$ to analyze the effectiveness of the key component in BST. The results are shown in Table \ref{table:aba} and Fig. \ref{fig:aba_selection} and discussed in detail below.

\noindent\textbf{Width Slimming.}
In Table \ref{table:aba_width}, we show the results of the MCAN$_\texttt{BST}$ variants trained with different width slimming strategies, \emph{i.e.}, the \emph{slim-all} strategy introduced in this paper and the \emph{slim-intermediate} strategy introduced in \cite{hou2020dynabert}. By comparing two submodels of similar FLOPs, we see that our \emph{slim-all} strategy delivers better model performance than the \emph{slim-intermediate} strategy under different model depths.

\noindent\textbf{Depth Slimming.} In Table \ref{table:aba_depth}, we compare the MCAN$_\texttt{BST}$ variants with different depth slimming strategies (mentioned in Section \ref{sec:the_bst_framework}) in terms of average accuracy over the ten submodels. From the results, we can see that the \emph{slim-middle} strategy achieves the best performance among the counterparts, suggesting that the bottom and top layers of the Transformer are more important than the middle layers. This observation is consistent with the results in \cite{wang2020rethinking}. We also provide visualization results in section \ref{sec:vis} by introducing a simple score function based on the magnitudes of the output features of each layer. The calculated layer scores are consistent with those obtained by the slim-middle strategy, verifying its effectiveness from a side view.  

\noindent\textbf{Number of Sampled Submodels.} In Table \ref{table:aba_numsamples}, we ablate the effects of different numbers of sampled submodels during the BST training. The results suggest that $K$=4 is a good trade-off between the average accuracy and training time. Further increasing $K$ does not bring a great performance improvement but takes much more training time. The fast saturation with such a small $K$ is facilitated by the weight-sharing strategy of different submodels.

\noindent\textbf{Training Strategy.} Our default training strategy uses the model parameters from a teacher model for initialization and then trains the submodels using a KD training strategy to exploit the implicit knowledge from the teacher model. The results in Table \ref{table:aba_train} show that both the teacher model initialization and the KD training improve the obtained MCAN$_\texttt{BST}$ model, compared to the model variants trained with random initialization or supervised by the ground-truth answer. In contrast to our KD training strategy that uses a fixed teacher model, an alternative strategy introduces a special \emph{inplace distillation} (ID) training strategy that takes the largest submodel as the \emph{dynamic} teacher to perform knowledge distillation \cite{yu2019universally}. We note that the ID-based strategy results in worse performance than the KD training strategy (65.55\% \emph{vs.} 66.95\% in terms of average accuracy), and even underperforms standard training without knowledge distillation (65.55\% \emph{vs.} 65.89\%). This suggests that a stable teacher model plays a key role in BST.
\begin{figure}
	\begin{center}
		\includegraphics[width=\columnwidth]{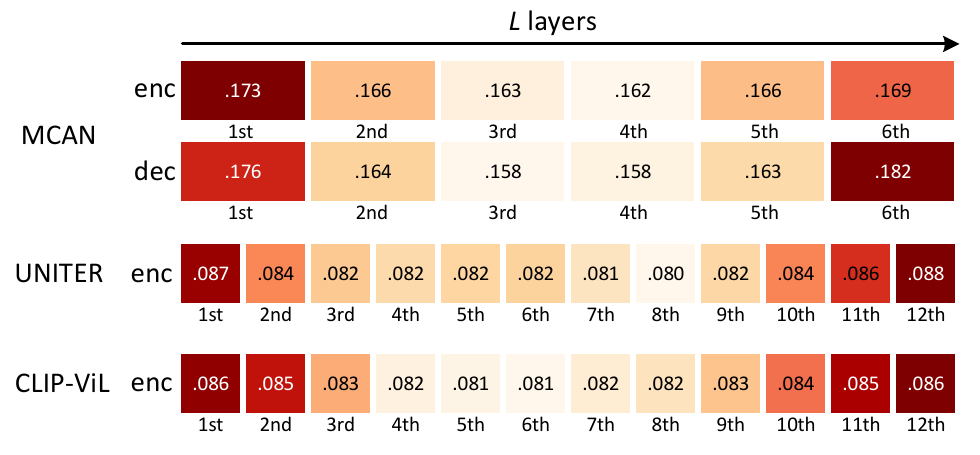}
		\caption{Visualizations of the calculated layer importance scores given trained MCAN, UNITER, and CLIP-ViL models, respectively. A darker color indicates a larger score of the layer. The layer scores for all the three models exhibit a ``heavy ends and light middle'' phenomenon, which are consistent with the layer scores achieved by the \emph{slim-middle} strategy.}
		\vspace{-15pt}
		\label{fig:layer_score}
	\end{center}
\end{figure}

\begin{figure*}
	\begin{center}
		\includegraphics[width=\textwidth]{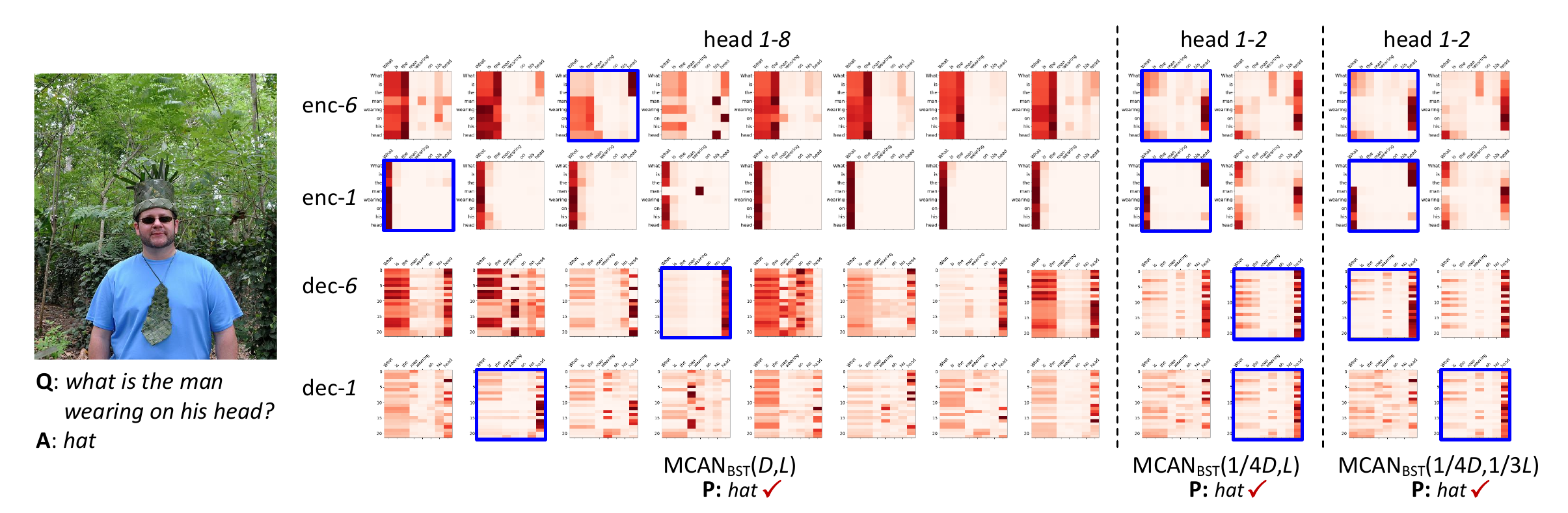}
		\caption{Visualizations of the learned attention maps ($\mathrm{softmax}(QK^T/\sqrt{D_H})$) from three MCAN$_\texttt{BST}$ submodels trained on VQA-v2. For each submodel, we show the attention maps from different heads in the first and last layers of the question encoder and image decoder, respectively. We highlight some representative attention maps with blue bounding boxes to better understand the attention mechnisims behind.}
		\vspace{-15pt}
		\label{fig:visualize}
	\end{center}
\end{figure*}

\noindent\textbf{Submodel Selection.} In Fig. \ref{fig:aba_selection}, we compare two MCAN$_\texttt{BST}$ variants with 10 and 16 submodels. From the results, we have the following observations: 1)  all the \emph{ineffective} submodels, which require more FLOPs but obtain lower accuracies than some other submodels, are precisely detected by our simple triangle selection strategy, 2) removing such ineffective submodel architectures before the BST training brings improvements to all the remaining submodels, and 3) taking the submodel (3/4$D$, 2/3$L$) as the reference model, the submodel (3/4$D$, $L$) outperforms ($D$, 2/3$L$) with fewer FLOPs, verifying our hypothesis that increasing depth is more economical than increasing width for the Transformer.

\subsection{Visualization Analysis}\label{sec:vis}
As discussed in section \ref{sec:aba}, the method of measuring the layer importance plays key role in depth slimming. 
Here, we introduce a data-driven strategy to measure the importance of each layer. Specifically, given a trained MCAN (UNITER or CLIP-ViL) model without model slimming, we first feed all the training samples through the network and memorize the output features for each layer. For the  output features obtained from each layer, mean pooling is performed on each flattened feature vector to obtain its unnormalized score. Using this strategy, we show the importance scores for each layer of the MCAN, UNITER, and CLIP-ViL models in Fig. \ref{fig:layer_score}. From the visualized results, we see that the layer scores of all models exhibit ``heavy tails and a light middle'', which is consistent with the layer scores achieved by the slim-middle strategy.

To better understand the behaviors of the weight-sharing submodels learned by BST, we show the attention maps from three MCAN$_\texttt{BST}$ submodels in Fig. \ref{fig:visualize}. Due to space limitations, we only show one example and visualize the attention maps from the first and last layers of the question encoder and image decoder, respectively. To better understand the effect of the attention mechanism, we highlight some representative attention maps with blue bounding boxes. From the results, we have the following observations. In general, the slimmed submodels MCAN$_\texttt{BST}$(1/4$D$,$L$) and MCAN$_\texttt{BST}$(1/4$D$,1/3$L$) have fewer \emph{redundant} heads (\emph{i.e.}, similar attention maps within one layer) than the full-sized model MCAN$_\texttt{BST}$($D$,$L$). This verifies the feasibility and necessity of our BST framework. Moreover, the three submodels, which have different widths and depths, all predict the correct answer. This verifies the effectiveness of both the width and depth slimming strategies, as well as the training paradigm. Furthermore, the attention maps from the different submodels have similar properties to the attention maps in the original MCAN paper \cite{yu2019mcan}. Almost all the attention maps from the first layer of the question encoder (\emph{i.e.}, enc-1) attend to the column of words such as `\emph{what}', which act as the question type classifiers. In contrast, some attention maps from the last layer of the question encoder (\emph{i.e.}, enc-6) and image decoder (\emph{i.e.}, dec-1 and dec-6) focus on the columns of keywords such as `\emph{head}'.

\section{Conclusion}
In this paper, we present a new direction for the VQA task: learning efficient and elastic models that can adaptively fit different platforms. To this end, we present a general bilaterally slimmable Transformer (BST) framework that can be seamlessly integrated with any Transformer-based VQA model in theory. By integrating the BST framework with three typical Transformer-based VQA approaches, the resulting slimmable models outperform state-of-the-art methods with similar model sizes, or achieve comparable performance to that of much smaller models on both VQA-v2 and GQA datasets. Moreover, the quantitative experiments on diverse hardware platforms and devices show the practicability and necessity of BST.

To the best of our knowledge, the proposed BST framework is the {first} attempt to explore efficient and elastic models for VQA. We hope our general framework can serve as a baseline to inspire future research on efficient multimodal learning.

\ifCLASSOPTIONcaptionsoff
  \newpage
\fi



\bibliographystyle{IEEEtran}
\bibliography{bst}





\end{document}